\documentclass[11pt,a4paper]{article}
\usepackage{times}

\usepackage{calc}

\usepackage{amsmath}
\usepackage{amssymb}
\usepackage{mathrsfs}
\usepackage{interval}

\usepackage[caption=false,font=footnotesize]{subfig}

\usepackage{graphicx}
\usepackage{booktabs}
\usepackage{makecell}
\usepackage{multirow}
\usepackage{amsfonts}
\usepackage{url}

\newlength{\Lpr}
\newsavebox{\Bpr}

\newcommand{\V}[1]{\mbox{\boldmath$#1$\unboldmath}}

\newcommand{\tr}{\ensuremath{{\mathrm{tr}}}}

\newcommand{\bdm}{\begin{displaymath}}
\newcommand{\edm}{\end{displaymath}}

\newcommand{\be}[1]{\begin{equation} \label{#1}}
\newcommand{\ee}{\end{equation}}

\newcommand{\bae}[3]{
\begin{equation} \label{#1}
\renewcommand{\arraystretch}{#2}
\begin{array}{#3}}

\newcommand{\eae}{\end{array}\end{equation}}

\newcommand{\baen}[2]{
\begin{displaymath} 
\renewcommand{\arraystretch}{#1}
\begin{array}{#2}}

\newcommand{\eaen}{\end{array}\end{displaymath}}

\newcommand{\DefLetter}[4]{
\newcommand{#1}{\ensuremath{\V{#2}}} 
\newcommand{#3}{\ensuremath{\V{#4}}} 
}

\DefLetter{\vzer}{0}{\mzer}{0}
\DefLetter{\vone}{1}{\mone}{1}
\DefLetter{\va}{a}{\ma}{A}
\DefLetter{\vb}{b}{\mb}{B}
\DefLetter{\vc}{c}{\mc}{C}
\DefLetter{\vd}{d}{\md}{D}
\DefLetter{\ve}{e}{\me}{E}
\DefLetter{\vf}{f}{\mf}{F}
\DefLetter{\vg}{g}{\mg}{G}
\DefLetter{\vh}{h}{\mh}{H}
\DefLetter{\vi}{i}{\mi}{I}
\DefLetter{\vj}{j}{\mj}{J}
\DefLetter{\vk}{k}{\mk}{K}
\DefLetter{\vl}{l}{\ml}{L}
\DefLetter{\vm}{m}{\mm}{M}
\DefLetter{\vn}{n}{\mn}{N}
\DefLetter{\vpr}{p}{\mpr}{P}
\DefLetter{\vq}{q}{\mq}{Q}
\DefLetter{\vr}{r}{\mr}{R}
\DefLetter{\vs}{s}{\ms}{S}
\DefLetter{\vt}{t}{\mt}{T}
\DefLetter{\vur}{u}{\mur}{U}
\DefLetter{\vv}{v}{\mv}{V}
\DefLetter{\vw}{w}{\mw}{W}
\DefLetter{\vx}{x}{\mx}{X}
\DefLetter{\vy}{y}{\my}{Y}
\DefLetter{\vz}{z}{\mz}{Z}

\DefLetter{\vdel}{\delta}{\mdel}{\Delta}
\DefLetter{\vphi}{\phi}{\mphi}{\Phi}
\DefLetter{\vpsi}{\psi}{\mpsi}{\Psi}
\DefLetter{\vrho}{\rho}{\mrho}{\Lambda}
\DefLetter{\vxi}{\xi}{\mxi}{\Xi}

\DefLetter{\valpha}{\alpha}{\malpha}{\Alpha}
\DefLetter{\vbeta}{\beta}{\mbeta}{\Beta}
\DefLetter{\vlam}{\lambda}{\mlam}{\Lambda}
\DefLetter{\vsig}{\sigma}{\msig}{\Sigma}
\DefLetter{\vtau}{\tau}{\mtau}{\tau}
\DefLetter{\vtheta}{\theta}{\mtheta}{\Theta}
\DefLetter{\vome}{\omega}{\mome}{\Omega}
\DefLetter{\vzero}{0}{\mzero}{0}
\DefLetter{\vgam}{\gamma}{\mgam}{\Gamma}
\DefLetter{\veps}{\epsilon}{\meps}{\Epsilon}
\DefLetter{\veta}{\eta}{\meta}{\Eta}

\newcommand{\DefFuncLetter}[2]{
\newcommand{#1}{\ensuremath{{#2}}} 
}

\DefFuncLetter{\Fzer}{0}
\DefFuncLetter{\Fa}{a}
\DefFuncLetter{\FA}{A}
\DefFuncLetter{\Fb}{b}
\DefFuncLetter{\Fc}{c}
\DefFuncLetter{\FC}{C}
\DefFuncLetter{\Fd}{d}
\DefFuncLetter{\Fe}{e}
\DefFuncLetter{\Ff}{f}
\DefFuncLetter{\Fg}{g}
\DefFuncLetter{\FG}{G}
\DefFuncLetter{\Fh}{h}
\DefFuncLetter{\FH}{H}
\DefFuncLetter{\Fi}{i}
\DefFuncLetter{\Fk}{k}
\DefFuncLetter{\Fl}{l}
\DefFuncLetter{\FL}{L}
\DefFuncLetter{\Fm}{m}
\DefFuncLetter{\Fn}{n}
\DefFuncLetter{\Fnr}{n}
\DefFuncLetter{\FN}{N}
\DefFuncLetter{\Fo}{o}
\DefFuncLetter{\FO}{O}
\DefFuncLetter{\Fpr}{p}
\DefFuncLetter{\FPr}{P}
\DefFuncLetter{\Fq}{q}
\DefFuncLetter{\Fr}{r}
\DefFuncLetter{\Fs}{s}
\DefFuncLetter{\FS}{S}
\DefFuncLetter{\Ft}{t}
\DefFuncLetter{\FT}{T}
\DefFuncLetter{\Fu}{u}
\DefFuncLetter{\FU}{U}
\DefFuncLetter{\Fv}{v}
\DefFuncLetter{\Fw}{w}
\DefFuncLetter{\FW}{W}
\DefFuncLetter{\Fx}{x}
\DefFuncLetter{\Fy}{y}
\DefFuncLetter{\FY}{Y}
\DefFuncLetter{\Fz}{z}
\DefFuncLetter{\FZ}{Z}
\DefFuncLetter{\Falp}{\alpha}
\DefFuncLetter{\Fbet}{\beta}
\DefFuncLetter{\Fchi}{\chi}
\DefFuncLetter{\Fdel}{\delta}
\DefFuncLetter{\Fzet}{\zeta}
\DefFuncLetter{\FEps}{\Epsilon}
\DefFuncLetter{\Feta}{\eta}
\DefFuncLetter{\Fphi}{\phi}
\DefFuncLetter{\FPhi}{\Phi}
\DefFuncLetter{\Fpsi}{\psi}
\DefFuncLetter{\FPsi}{\Psi}
\DefFuncLetter{\Fgam}{\gamma}
\DefFuncLetter{\FGam}{\Gamma}
\DefFuncLetter{\Flam}{\lambda}
\DefFuncLetter{\FLam}{\Lambda}
\DefFuncLetter{\Fsig}{\sigma}
\DefFuncLetter{\Ftau}{\tau}
\DefFuncLetter{\Fome}{\omega}
\DefFuncLetter{\Feps}{\epsilon}
\DefFuncLetter{\Fthe}{\theta}
\DefFuncLetter{\Fvar}{\vartheta}

\DefFuncLetter{\FB}{B}
\DefFuncLetter{\FD}{D}
\DefFuncLetter{\FE}{E}
\DefFuncLetter{\FF}{F}
\DefFuncLetter{\FI}{I}
\DefFuncLetter{\FJ}{J}
\DefFuncLetter{\FM}{M}
\DefFuncLetter{\FR}{R}
\DefFuncLetter{\FV}{V}
\DefFuncLetter{\FX}{X}

\newcommand{\DefCalLetter}[2]{
\newcommand{#1}{\ensuremath{\mathcal{#2}}} 
}
\DefCalLetter{\CC}{C}
\DefCalLetter{\CD}{D}

\DefCalLetter{\CS}{S}
\DefCalLetter{\CV}{V}

\newcommand{\DefSubLetter}[2]{
\newcommand{#1}{\mathrm{#2}} 
}

\DefSubLetter{\slzer}{0}
\DefSubLetter{\sla}{a}
\DefSubLetter{\slA}{A}
\DefSubLetter{\slb}{b}
\DefSubLetter{\slB}{B}
\DefSubLetter{\slc}{c}
\DefSubLetter{\slC}{C}
\DefSubLetter{\sld}{d}
\DefSubLetter{\slD}{D}
\DefSubLetter{\sle}{e}
\DefSubLetter{\slE}{E}
\DefSubLetter{\slf}{f}
\DefSubLetter{\slF}{F}
\DefSubLetter{\slg}{g}
\DefSubLetter{\slG}{G}
\DefSubLetter{\slh}{h}
\DefSubLetter{\slH}{H}
\DefSubLetter{\sli}{i}
\DefSubLetter{\slI}{I}
\DefSubLetter{\slk}{k}
\DefSubLetter{\sll}{l}
\DefSubLetter{\slL}{L}
\DefSubLetter{\slm}{m}
\DefSubLetter{\slM}{M}
\DefSubLetter{\sln}{n}
\DefSubLetter{\slnr}{n}
\DefSubLetter{\slN}{N}
\DefSubLetter{\slo}{o}
\DefSubLetter{\slp}{p}
\DefSubLetter{\slP}{P}
\DefSubLetter{\slq}{q}
\DefSubLetter{\slQ}{Q}
\DefSubLetter{\slr}{r}
\DefSubLetter{\slR}{R}
\DefSubLetter{\sls}{s}
\DefSubLetter{\slS}{S}
\DefSubLetter{\slt}{t}
\DefSubLetter{\slT}{T}
\DefSubLetter{\slu}{u}
\DefSubLetter{\slU}{U}
\DefSubLetter{\slv}{v}
\DefSubLetter{\slw}{w}
\DefSubLetter{\slW}{W}
\DefSubLetter{\slx}{x}
\DefSubLetter{\slX}{X}
\DefSubLetter{\sly}{y}
\DefSubLetter{\slY}{Y}
\DefSubLetter{\slz}{z}
\DefSubLetter{\slZ}{Z}

\DefSubLetter{\slalp}{\alpha}
\DefSubLetter{\slbet}{\beta}
\DefSubLetter{\sldel}{\delta}
\DefSubLetter{\slDel}{\Delta}
\DefSubLetter{\sleps}{\epsilon}
\DefSubLetter{\slgam}{\gamma}
\DefSubLetter{\slphi}{\phi}
\DefSubLetter{\sltau}{\tau}
\DefSubLetter{\slxi}{\xi}
\DefSubLetter{\slthe}{\theta}

\newcommand{\ra}[1]{\renewcommand{\arraystretch}{#1}}

\usepackage{color}

\begin{document}
	%
\begin{center}
\parbox{\textwidth}{\hrulefill}\\ 
\textbf{\Large
Deep Learning Based Speckle Filtering for Polarimetric SAR Images.\\ Application to Sentinel-1}\\
\parbox{\textwidth}{\hrulefill}\\
\textbf{\Large A preprint}\\
\parbox{\textwidth}{\hrulefill}\\

Alejandro~Mestre-Quereda and
		Juan~M.~Lopez-Sanchez\footnote{This work was supported by the the European Funds for Regional Development and by the Spanish Ministry of Science and Innovation (Agencia Estatal de Investigaci\'on, AEI) with Project PID2020-117303GB-C22/AEI/10.13039/501100011033.\\A.~Mestre-Quereda and J.~M.~Lopez-Sanchez 
			are with the Signals, Systems and Telecommunications Group, Institute for Computer Research (IUII), University
			of Alicante, Alicante, Spain (e-mail: alejandro.mestre@ua.es, juanma-lopez@ieee.org).}   
	
\end{center}	
	
	
	
	\begin{abstract}
		
	Speckle suppression in synthetic aperture radar (SAR) images is a key processing step which continues to be a research topic. A wide variety of methods, using either spatially-based approaches or transform-based strategies, have been developed and have shown to provide outstanding results. However, recent advances in deep learning techniques and their application to SAR image despeckling have been demonstrated to offer state-of-the-art results. Unfortunately, they have been mostly applied to single-polarimetric images. The extension of a deep learning-based approach for speckle removal to polarimetric SAR (PolSAR) images is complicated because of the complex nature of the measured covariance matrices for every image pixel, the properties of which must be preserved during filtering. In this work, we propose a complete framework to remove speckle in polarimetric SAR images using a convolutional neural network. The methodology includes a reversible transformation of the original complex covariance matrix to obtain a set of real-valued intensity bands which are fed to the neural network. In addition, the proposed method includes a change detection strategy to avoid the neural network to learn erroneous features in areas strongly affected by temporal changes, so that the network only learns the underlying speckle component present in the data. The method is implemented and tested with dual-polarimetric images acquired by Sentinel-1. Experiments show that the proposed approach offers exceptional results in both speckle reduction and resolution preservation. More importantly, it is also shown that the neural network is not generating artifacts or introducing bias in the filtered images, making them suitable for further polarimetric processing and exploitation. 
	
	\end{abstract}
	

	%

	\section{Introduction}
	Synthetic Aperture Radars (SAR) constitute nowadays relevant instruments for Earth observation, due to their wide coverage, weather-independent capabilities, and because they can acquire images indistinctly during day and night.
	Conventional SAR systems were designed to gather images using a single combination of polarizations of transmitted and received electromagnetic signals, hence denoted as single-pol images. However, modern SAR sensors are able to acquire data in multiple polarization combinations. SAR Polarimetry (abbreviated PolSAR)~\cite{cloudeBookPolarisation2009, lee2009polarimetric} exploits this polarization diversity to provide more information content about the properties of the imaged area. PolSAR is applied in a wide variety of domains, including forestry, agriculture, wetlands, oceans, ice, and urban studies~\cite{b:polsarapp}.
	
	One of the problems of the analysis of SAR images is the so-called \textit{speckle}~\cite{Lee1981, Curlander1991}, caused by the constructive or destructive interferences of signals from multiple scatterers within a SAR resolution cell. Even though, strictly speaking, speckle is not noise, it is usually considered as such as a consequence of the granular aspect that it gives to the images, which degrades their quality and hampers their interpretability and further exploitation. Since the speckle component of a SAR image cannot be analytically predicted, it is considered as a noisy component present in the measured data.
	
	Speckle reduction or \textit{despeckling} can be achieved by signal processing or filtering techniques. The simplest filter is the so-called multilook, which, basically, consists in averaging independent samples (or looks) of the SAR image. It is commonly applied in the spatial domain using a pre-defined window of a given size. Multilooking effectively smooths homogeneous areas by reducing the variance due to speckle, but inevitably causes a loss of spatial resolution and blurs fine details present in the image. Increasing the window size leads to a better speckle suppression, but also to a stronger degradation of the  spatial resolution caused by the average of heterogeneous areas (i.e., the average of different types of surfaces or targets). To limit this loss of resolution, many \textit{spatial-based} adaptive methods using local statistics were proposed, e.g.~\cite{Lee1980, Frost1982, Kuan1985, Lee1983, Lopes1990, Vasile2006}. State-of-the-art filters based on this strategy that provide outstanding results are Block-Matching 3-D (BM3D)~\cite{BM3D_or} (the filter was not originally proposed for SAR despeckling but was easily adapted), Non-Local SAR (NL-SAR)~\cite{Deledalle2015a}, and MUlti-channel LOgarithm with Gaussian denoising (MuLoG)~\cite{Deledalle_MuLoG}. Besides spatial filtering,  alternative strategies for speckle reduction using \textit{transform-based} approaches have also been proposed. The core idea of these methods is to separate the speckle component from the useful part of the signal by means of a domain transformation. Wavelet-based filters~\cite{Solbo2008,Gao_wavelets} and Total-Variational (TV) methods~\cite{Nie2016_TV} constitute examples of such transform-based approaches. 
	
	In the last years, the use of Deep Learning and Artificial Intelligence techniques with SAR data has experienced a significant growth~\cite{Xiao2021}. Deep learning-based methods represent a paradigm change: rather than on pre-defined signal processing algorithms or theoretical models, these methods depend only on the data. In this regard, speckle filtering with deep learning has shown to be not only possible, but able to provide results similar to the best, state-of-the-art SAR speckle filters based on conventional methods. Different works have shown how to successfully train Convolutional Neural Networks (CNN) for speckle suppression~\cite{Wang2017, Zhang2018, Lattari2019, Dalsasso2020, Sebastianelli2022, Dalsasso2021}. 
 
	
	All the previously mentioned deep learning-based methods are designed to remove speckle from the intensity of single-pol SAR images, i.e., the backscattering coefficient of a polarimetric channel. Unfortunately, their extension to PolSAR data is not straightforward. This is mainly due to the nature of the polarimetric information, which involves not only intensities but also correlations and phase differences between channels. Polarimetric data are expressed in the form of a complex covariance matrix. For each image pixel, this is an Hermitian, positive semi-definite matrix whose properties must be preserved independently of the filtering technique, so that further polarimetric processing can be properly achieved. To guarantee this preservation, all entries of the matrix must be filtered in exactly the same way~\cite{lee2009polarimetric}.
	
	To date, the adaption of deep learning-based speckle filters to polarimetric SAR images has not been fully evaluated, and few works are found in the literature. The framework proposed in~\cite{Tucker2022} adapts a residual neural network to fully-polarimetric data but uses a previously multilooked (spatially averaged) version of the original covariance matrix as reference. Although the methodology successfully removes speckle both on simulated and real PolSAR data, the spatial resolution of the filtered images is inevitably degraded because the network is not trained with full-resolution data. Moreover, in~\cite{Mulissa2022} a complex-valued neural network is trained for speckle reduction. This is arguably the most intuitive approach for despeckling PolSAR images (since the covariance matrix is complex), but the synthesis of an adequate reference (speckle-free) complex covariance matrix is almost impossible to achieve. Also, note that the method in~\cite{Mulissa2022} uses an additional filtering (MuLoG) before training, so it is not completely based on deep learning. An alternative strategy for despeckling polarimetric images is proposed in~\cite{Lin2023}. In this work, instead of training a neural network to estimate the filtered pixel values or the speckle component to be removed from the data, the network is trained to predict the weight matrices of the neighbors pixels of the pixel under test (i.e., the network finds similar pixels to average with the central one). This approach is motivated because of the difficulty of generating adequate reference (speckle-free) polarimetric images to train a model. As shown in~\cite{Lin2023}, the despeckling framework provides outstanding results, but again it is not completely based on deep learning because reference data are generated by prefiltering polarimetric images using conventional methods. 
 
    It is therefore evident that the major difficulty of using deep learning to remove speckle in polarimetric images is the lack of reference data and the
    generation of an adequate training dataset. As detailed in~\cite{Lu2024}, the exploitation of multi-temporal stacks, the generation of simulated speckled images or the combination of both strategies are the only possible ways of building training datasets, each one having different pros and cons.
	
	In this work, a complete, end-to-end  speckle filtering framework based on convolutional neural networks is proposed for polarimetric images. Due to its relevance in many applications and its open access policy, the method is evaluated using Sentinel-1 dual-pol images. The filtering process is based on the training of a well-known residual denoising convolutional neural network DnCNN~\cite{DnCNN}, for which a transformation of the complex polarimetric covariance matrix into a set of intensities is used. The neural network is trained using only real SAR images and includes a change detection strategy. As a consequence, the model will learn the spatial distribution of speckle and its fluctuations only from real data, so that it will be able to properly generalize to remove it in unknown (unseen) images.
	The paper is organized as follows. Section~\ref{sec:theory} provides a review of PolSAR data and the required transformations that have to be done from the polarimetric covariance matrix to train the neural network. A quick revision of change detection theory applied to time series or stacks of SAR images is also included in Section~\ref{sec:theory}. Then, Section~\ref{sec:method} describes all details of the proposed method, including a description of the network architecture, the procedure for data preparation, and an explanation of how the training dataset is generated. The evaluation of the performance of the proposed filter is presented in Section~\ref{sec:results}. Finally, main conclusions are summarized in Section~\ref{sec:concl}

	\section{Theoretical Basis}
	\label{sec:theory}
	\subsection{Polarimetric Covariance Matrix}
	\label{ssec:polsar_theory}
	
	Polarimetric SAR sensors measure different combinations of transmitting and receiving polarizations. The data are represented by means of the complex Hermitian positive semi-definite covariance matrix~\cite{lee2009polarimetric}, which size depends on the number of polarimetric channels measured. In dual-pol systems like Sentinel-1, the covariance matrix is denoted as $\mathbf{C}_2 \in \mathbf{Mat}_2(\mathbb{C})$. Assuming that the polarimetric channels are VV and VH, which are the most common over land for Sentinel-1, the single-look complex SAR images of the two polarimetric channels are denoted by $S_{\text{VV}}$ and $S_{\text{VH}}$, respectively. From them, the estimated covariance matrix is: 
	%
	\begin{equation}
			\label{eq:C2}
			\mathbf{C}_{2} = 
			\begin{bmatrix}
				C_{11} & C_{12}  \\
				C_{21} & C_{22}  \\
			\end{bmatrix} =
			\begin{bmatrix}
					E\{S_{\text{VV}}S^*_{\text{VV}}\}  & E\{S_{\text{VV}}S^*_{\text{VH}}\}  \\
					E\{S_{\text{VH}}S^*_{\text{VV}}\}   & E\{S_{\text{VH}}S^*_{\text{VH}}\}   \\
				\end{bmatrix}.
		\end{equation}
	where $^*$ denotes the complex conjugate, and $E\{\cdot\}$ refers to the mathematical expectation which, in practice, is substituted by a filtering operation.
	While the diagonal entries of $\mathbf{C}_2$ contain the intensities of each individual polarimetric channel (i.e., the backscattering coefficients), the off-diagonal entries represent the correlation between both channels. Note that these off-diagonal elements are complex numbers. 
 

	In this work, we have followed a supervised learning approach in which a CNN is trained on a high number of pairs of noisy (corrupted by speckle) and clean (reference) images, so that the network model ideally learns to remove the speckle. Evidently, speckle-free SAR observations cannot be obtained, so the reference image is formed as a multi-temporal average of a stack of images. 
 
 If the scene is stationary, i.e., there are no changes between the acquisitions, averaging the intensity of a multi-temporal stack provides a new intensity image that can be used as reference. However, in the case of polarimetric images, the temporal average of the covariance matrix does not provide a suitable reference which can be used to train neural networks. This is because, as shown in Equation~\ref{eq:C2}, the elements of the covariance matrix $\mathbf{C}_2$ contain different types of information (intensities in the diagonal elements and correlations in the off-diagonal entries) which follow different spatial and temporal statistics~\cite{CarlosLopez2003}.
	
 Therefore, to properly process polarimetric images, we have followed the same approach originally used in~\cite{Hoekman2003, Hoekman2011} with fully-polarimetric data, in which the elements of the original complex covariance matrix are transformed into a set of (positive) real-valued intensities. In the dual-pol case, we can define a bijective transformation $\Gamma (\cdot) : \mathbb{C}^{2\times2} \rightarrow \mathbb{R}^{4}_{+} $ which allows us to represent the polarimetric covariance matrix with four intensities:
	\begin{equation}
		\label{eq:C2_mod}
		\begin{cases} 
			\hat{c}_{vv} = E\{|S_{\text{VV}}|^2 \}\\
			\hat{c}_{i} =  E\{|S_{\text{VV}} + S_{\text{VH}}|^2 \}\\
			\hat{c}_{q} =  E\{|S_{\text{VH}} + jS_{\text{VV}}|^2 \}\\
			\hat{c}_{vh} =  E\{|S_{\text{VH}}|^2\},
		\end{cases}
	\end{equation}
 where $j=\sqrt{-1}$.
	%
	%
	%
	
 After filtering the four intensity bands, the inverse transformation $\Gamma^{-1} (\cdot) : \mathbb{R}^{4}_{+} \rightarrow \mathbb{C}^{2\times2} $ is applied to recover the  complex entries of the covariance matrix in its original form. The inverse transformation is explicitly expressed as follows:
	\begin{equation}
		\label{eq:C2_inv}
		\begin{cases} 
			C_{11} = \hat{c}_{vv} \\
			C_{12} = 0.5\cdot[ (\hat{c}_{i} - \text{SPAN}) + j(\hat{c}_{q} - \text{SPAN}) ] \\
			C_{21} = 0.5\cdot[ (\hat{c}_{i} - \text{SPAN}) - j(\hat{c}_{q} - \text{SPAN}) ] \\
			C_{22} = \hat{c}_{vh},
		\end{cases}
	\end{equation}
	where $\text{SPAN}$ is the total backscattered power corresponding to the summation of the intensities of both polarimetric channels (hence the trace of $\mathbf{C}_2$) :
	\begin{equation}
		\text{SPAN} = E\{|S_{\text{VV}}|^2 \} + E\{|S_{\text{VH}}|^2\}.
	\end{equation}
	Note that, essentially, we modify the way the polarimetric information is expressed to work with four intensity bands which fully describe the polarimetric information content of every image pixel. The equivalent transformation for fully-polarimetric data was presented in~\cite{Hoekman2003, Hoekman2011}. Finally, we note that the transformation employed here is not the same as the one used in~\cite{Tucker2022}. In our work all the bands are intensities (and thus positive real values) but the real and imaginary parts of the off-diagonal element of $\mathbf{C}_2$ were used in~\cite{Tucker2022}, which can be positive and negative and show statistics different from an intensity.    
	
	\subsection{Change Detection}
	\label{ssec:cd_theory}
	
	Averaging a large multi-temporal stack of co-registered SAR images  produces a new image with reduced speckle and where the original spatial resolution is effectively preserved. However, the application of a temporal averaging in areas that exhibit changes along time (such as agricultural areas) results in an image that is not representative of individual images acquired at a particular date and, therefore, cannot be used a reference for supervised learning. Therefore, the inclusion of a change detection strategy during the training process of the neural network is mandatory. The problem of temporal changes for despeckling SAR images with deep learning is still to be fully addressed and is usually not properly detailed. In fact, only the SAR2SAR algorithm~\cite{Dalsasso2021} describes a strategy for mitigating the impact of temporal changes, and it has shown to improve the filtering performance of the neural network by reducing the presence of bias/artifacts and oversmoothing effects. However, because the SAR2SAR follows a \emph{self-supervised} strategy based on the \textit{noise2noise} approach~\cite{noise2noise}, in which only noisy pairs are fed to the network (instead of pairs of noisy and clean images), temporal changes in the multi-temporal stack could only be partially compensated~\cite{Dalsasso2021}.
	
	In this work, a simple yet effective strategy to address the problem of temporal changes is proposed. Specifically, the well-known Omnibus Test~\cite{Conradsen_OmnibusTest} for detecting changes in time series of single or multi-polarimetric SAR data is used to discard image areas that significantly changed during the creation of the training dataset. The test evaluates the equality of $k$ covariance matrices ($k \geq 2$) of polarimetric SAR images. Each multilooked covariance matrix $\mathbf{C_i} = n\mathbf{\Sigma_i}$ of a PolSAR image acquired at time point $i$, follows a complex Wishart distribution~\cite{Conradsen_OmnibusTest, Nielsen2015}:
	\begin{equation}
		\mathbf{C_i}  \thicksim W_{C}(p, n, \mathbf{\Sigma_i}), i = 1, ..., k,
	\end{equation}
	where $p = 2$ for dual-pol data, and $n$ is the equivalent number of looks.
	The omnibus test providing potential changes in a stack of polarimetric images is:
	\begin{equation}
		Q = \biggl(k^{pk}\frac{\prod_{i=1}^{k}|\mathbf{C_i}| }{|\mathbf{C}|^k}\biggl)^n. 
	\end{equation} 
	where $|\cdot|$ denotes the determinant, $Q \in [0,1]$ with $Q = 0$ for inequality (total change), $Q = 1$ for equality (no change) of all covariance matrices of the dataset, and $\mathbf{C} = \sum_{i=1}^{k}\mathbf{C_i}$, $\mathbf{C}  \thicksim W_{C}(p, n, \mathbf{\Sigma_i})$.
	
	For the the natural logarithm of $Q$ we get~\cite{Conradsen_OmnibusTest}:
	\begin{equation}
		\ln Q = n\biggl(pk\ln k + \sum_{i=1}^{k}\ln\mathbf{C_i} - k\ln|\mathbf{C}| \biggl),
	\end{equation} 
	and the probability of finding a value of $-2\rho\ln Q$ smaller than an observed value $z$ ($z = -2\rho\ln q_{obs}$) is:
	\begin{multline}
		\label{eq:plnQ}
		P(-2\rho\ln Q \leq z) \simeq P(\chi^2(f) \leq z ) + \\
		\omega_2\cdot[P(\chi^2(f+4) \leq z) - P(\chi^2(f) \leq z) ]
	\end{multline}
	where $\chi^2(f)$ is the chi square probability distribution with $f$ degrees of freedom, and $\rho$, $f$ and $\omega_2$ depend on the number of images $k$, polarimetric channels $p$ (2 in this case) and the equivalent number of looks $n$:
	\begin{align*} 
		f &=  (k-1)p^2  \\ 
		\rho &=  1-\frac{(2p^2-1)}{6(k-1)p} \biggl( \frac{k}{n} - \frac{1}{nk}  \biggl) \\
		\omega_2 &=  \frac{p^2(p^2-1)}{24p^2} \biggl( \frac{k}{n^2} - \frac{1}{(nk)^2}  \biggl) - \frac{p^2(k-1)}{4} \biggl( 1- \frac{1}{\rho}  \biggl)^2.
	\end{align*}
	
	Finally, to decide whether the omnibus test detects a true change in the set of covariance matrices or just random fluctuations, a threshold is applied to the change probability $P(-2\rho\ln Q \leq z)$ (or the no-change probability $1 - P(-2\rho\ln Q \leq z)$) according to a given significance level.

	\section{Proposed Method}
	\label{sec:method}
	\subsection{General Description}
\begin{figure*}[h!]
		\centering
		\includegraphics[width = 1.0\textwidth]{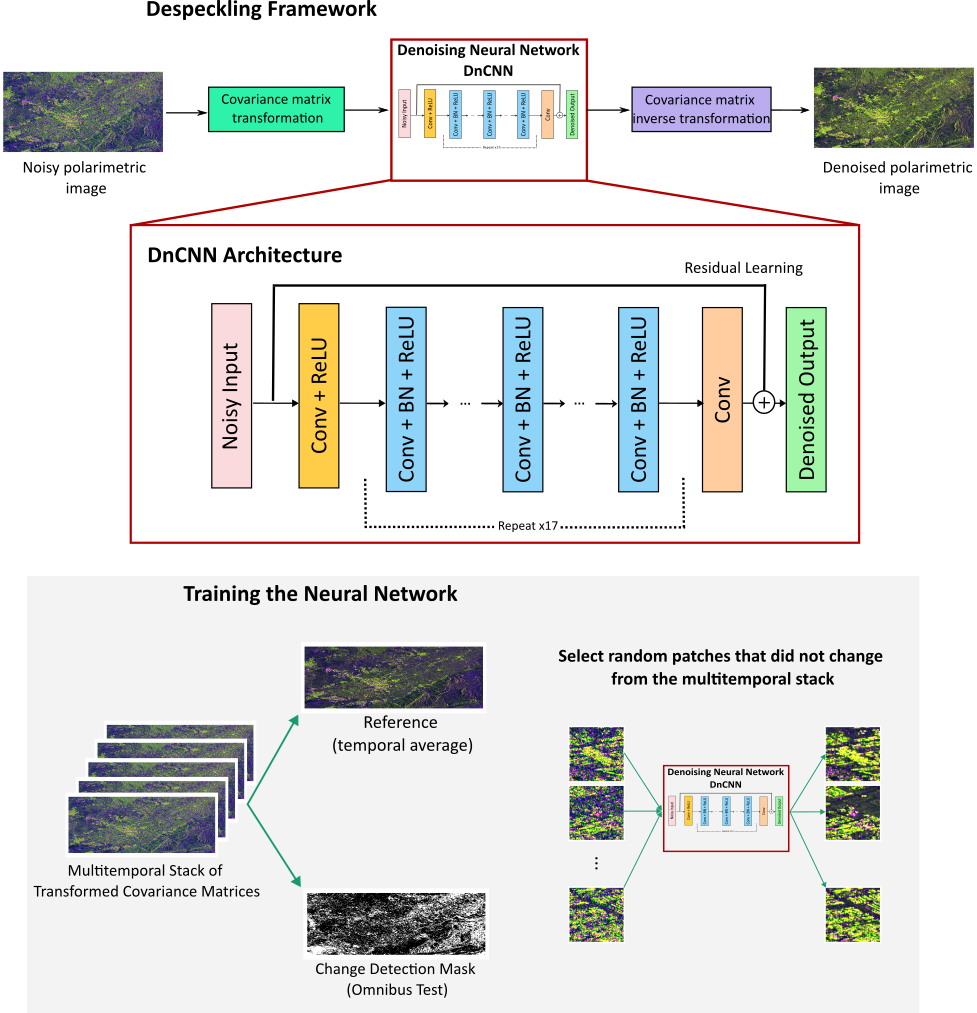}
		\caption{General scheme of the proposed polarimetric speckle filter based on the DnCNN neural network architecture.}
		\label{fig:polsar_cnn_scheme}
	\end{figure*}

	The general scheme of the proposed method is represented in Figure~\ref{fig:polsar_cnn_scheme}. The upper part of the scheme summarizes the despeckling process: the original input PolSAR image in the form of a complex covariance matrix is transformed into four real-valued intensity bands. Then, each band is despeckled by the deep neural network. Note that, unlike other despeckling algorithms based on deep learning that apply a logarithmic transformation of the input data~\cite{Dalsasso2021, Tucker2022}, the restoration here is performed in the linear domain. We preferred to keep the data in the original linear scale, in which conventional despeckling algorithms work, to avoid changing the statistics of the data. 
 Once the four intensities are despeckled, the complex covariance matrix form is recovered by means of the inverse transformation detailed in Section~\ref{ssec:polsar_theory}. The central part of Fig.~\ref{fig:polsar_cnn_scheme} shows the network architecture of the DnCNN (\textit{Denoising Convolutional Neural Network}). Originally proposed for Gaussian denoising in~\cite{DnCNN}, it follows a residual learning approach in which, instead of outputting directly a denoised image, the network is trained to estimate a residual image that is the difference between the noisy (measured) observation and the latent noise-free image. The lower part of Fig.~\ref{fig:polsar_cnn_scheme} shows how the neural network is trained using supervised learning: a set of time series of polarimetric data are temporally averaged to create a set of reference images, and a large number of small patches are randomly selected to create the training dataset. For each temporal stack, a change detection mask is calculated using the omnibus test explained in Section~\ref{ssec:cd_theory}. Note that Fig.~\ref{fig:polsar_cnn_scheme} only shows one  multi-temporal stack, but, in practice, we have employed multiple stacks acquired in different sites as it will be detailed in Section~\ref{ssec:dataset}.

	\subsection{Neural Network Architecture}
	We have used the original architecture of DnCNN proposed in~\cite{DnCNN}, which is basically composed of three types of layers as shown in Fig.~\ref{fig:polsar_cnn_scheme}. A first layer Conv+ReLU composed of 64 convolutional filters of size $3\times3\times4$ (4 corresponds to the number of bands) is used to generate 64 feature maps, followed by a Rectified Linear Unit (ReLU) for nonlinearity. The next 17 layers Conv+BN+ReLU are composed of 64 filters of size $3\times3\times64$, and batch normalization~\cite{BN} is added between the convolution layer and ReLU. The last layer Conv uses 4 filters of size $3\times3\times4$ to reconstruct the output of the network. 
	
	Moreover, as well as in~\cite{DnCNN} or in the SAR2SAR approach~\cite{Dalsasso2021}, the residual learning approach aims to estimate the noise $\mathbf{v}$ of the images $\mathbf{y}$ as the residual $R(\mathbf{y}) \approx \mathbf{v}$. This means that, in our case, the CNN tries to estimate the speckle component, which is then removed from the original images. The sum squared error between the residual of the reference images and the estimated ones with the neural network is used as the loss function:
	\begin{equation}
		\boldsymbol\ell(\boldsymbol{\Theta})  = \sum_{i=1}^{N}|| R(\mathbf{y}; \Theta) - (\mathbf{y_i} - \mathbf{x_i})||^2,
	\end{equation} 
	where $\boldsymbol\Theta$ is the set of trainable parameters of the network, and $\{(\mathbf{y_i} - \mathbf{x_i})\}_{i=1}^N$ are $N$ image patches obtained as differences between the noisy $\mathbf{y_i}$ and the clean $\mathbf{x_i}$ training images (hereafter denoted as noisy--clean patches).
	
	\subsection{Data Normalization}
	
	Data normalization is a crucial step that must be taken into account to facilitate the neural network optimization process. In this case, we have followed a conventional $\min/\max$ (linear) normalization of the four intensities according to:
	\begin{equation}
		\label{eq:norm}
		\mathbf{X_{N}} = \frac{\mathbf{X} - x_{\min}}{x_{\max}  - x_{\min}},
	\end{equation}
	where $\mathbf{X}$ is the original intensity, and $x_{min}$ and $x_{max}$ represent the minimum and maximum values, respectively, which allow to scale the data in the range $[0, 1]$.
	Note that the output intensity bands are rescaled to their original dynamic range by means of:
	\begin{equation}
		\mathbf{X} = \mathbf{X_{N}} \cdot (x_{\max} - x_{\min} ) + x_{\min}
	\end{equation}

	\subsection{Training Dataset Generation}
	\label{ssec:dataset}
	
	We have considered three long time series of dual-polarimetric SAR images, acquired by Sentinel-1 at VV and VH channels, over different areas and during three or four consecutive years. The nominal single-look spatial resolution of the images is approximately 3 m $\times$ 22 m in range and azimuth, respectively. All acquisitions of each time series were radiometrically calibrated and coregistered. More details of the dataset are shown in Table~\ref{t:dataset_train}.
	\begin{table*}[ht!]
	
	\centering
	\caption{Characteristics of Sentinel-1 polarimetric images used for training.}
	\begin{tabular}{@{}lccccc@{}}
		\toprule
		Geographic area & Orbit (type) & \thead{Processed image dimensions \\ azimuth $\times$ range}  & Acquisition dates & Number of images \\
		\toprule
		Murcia (Spain)  & 8 (Descending)  & 2389 $\times$ 5372   & 2017-01-05 to 2019-12-27   &    177 \\
		Toscana (Italy) & 117 (Ascending) & 10000 $\times$ 15000 & 2019-01-01 to 2021-12-22   & 175 \\
		Toscana (Italy) & 95 (Descending)  & 7000 $\times$ 10000  & 2019-01-02 to 2022-12-23   & 190  \\
		\bottomrule
	\end{tabular}
	\label{t:dataset_train}
	\end{table*}

\begin{figure*}
		\centering
		\setkeys{Gin}{width=0.9\linewidth}
		\subfloat[Image acquired at date 2017-04-05.]{\includegraphics{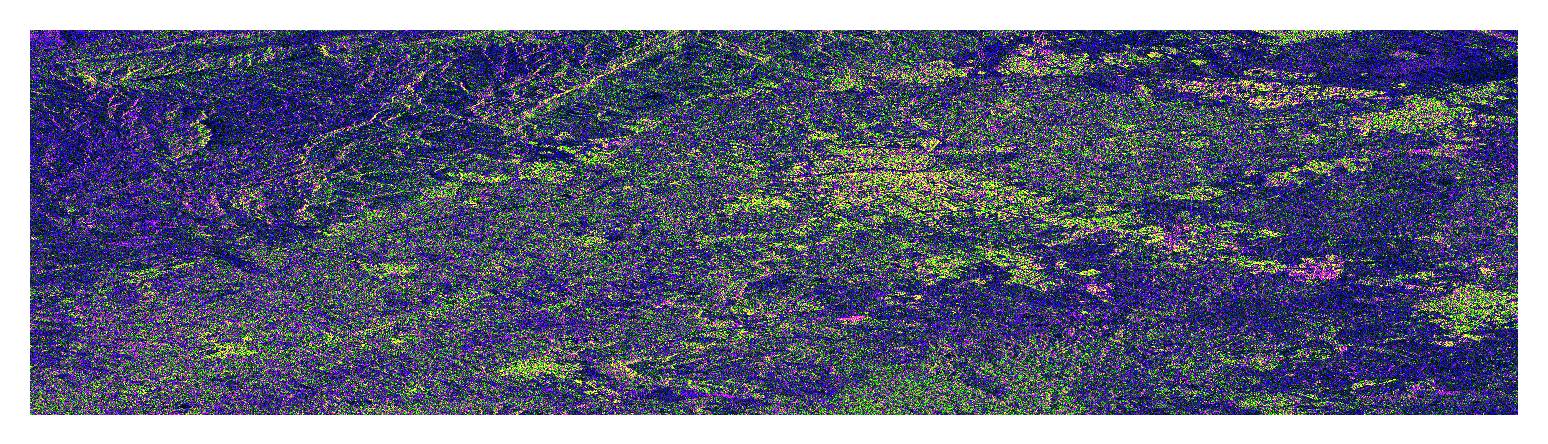}}  \\
		\subfloat[Temporal average.]{\includegraphics{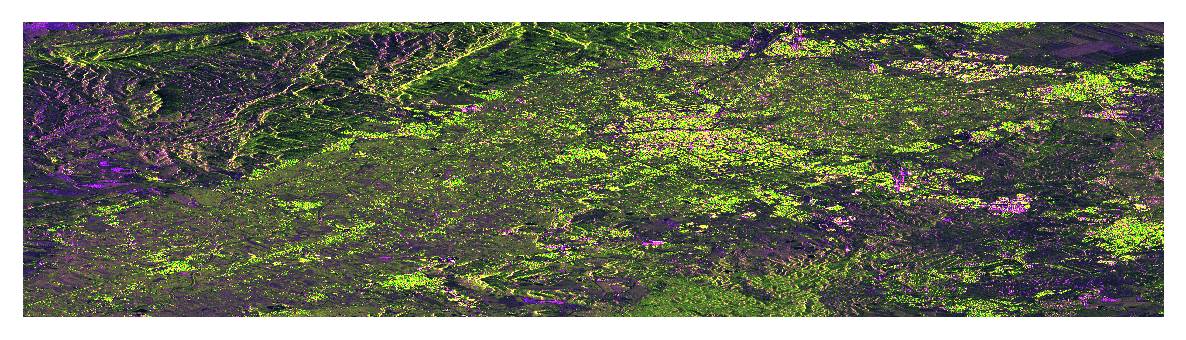}}
		
		\caption{Images of the dataset of Murcia (Spain): (a) single image at a randombly selected date, and (b) temporal average of the whole time series. Representation: RGB false colour composite formed with the backscattering coefficient at the linear channels with $R = VV$, $G = VH$, $B = VV/VH$.}
		\label{fig:reference_imgs}
	\end{figure*}

	As an example, Fig.~\ref{fig:reference_imgs} shows a false-color RGB representation of a randomly selected image of the time series of Murcia and also the temporal average. As it can be observed in Fig.~\ref{fig:reference_imgs}(b) speckle is considerably reduced by the temporal averaging and the spatial resolution is preserved.
	
\begin{figure*}
		\centering
		\setkeys{Gin}{width=0.5\linewidth}
		\subfloat[Reference image.]{\includegraphics{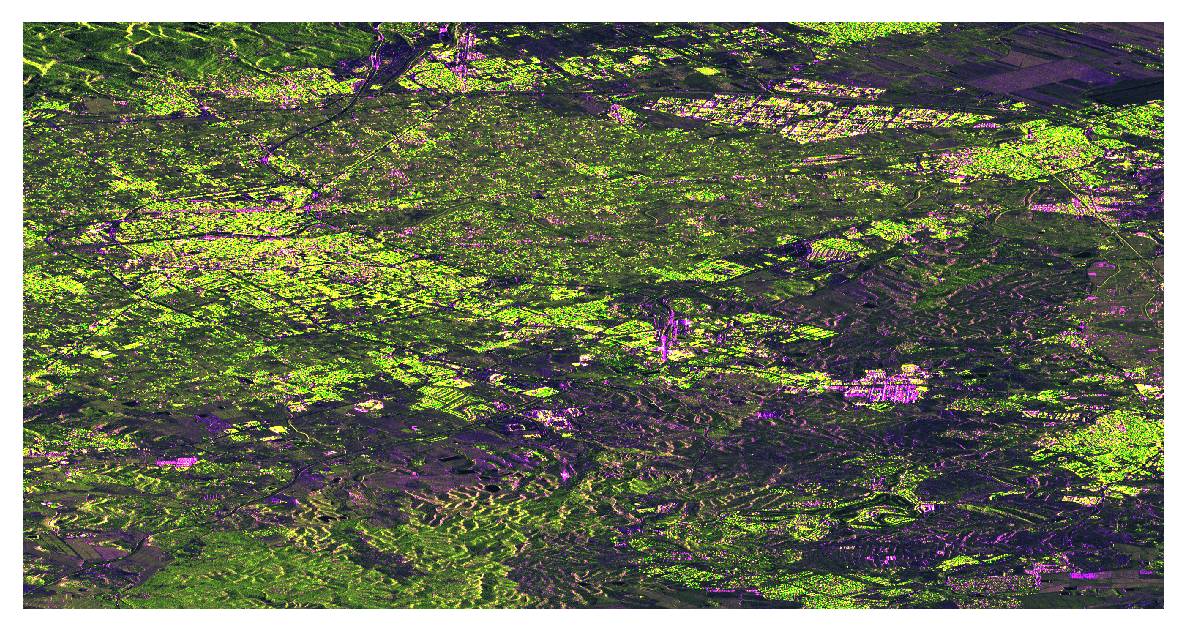}} 
		\subfloat[Change detection mask.]{\includegraphics{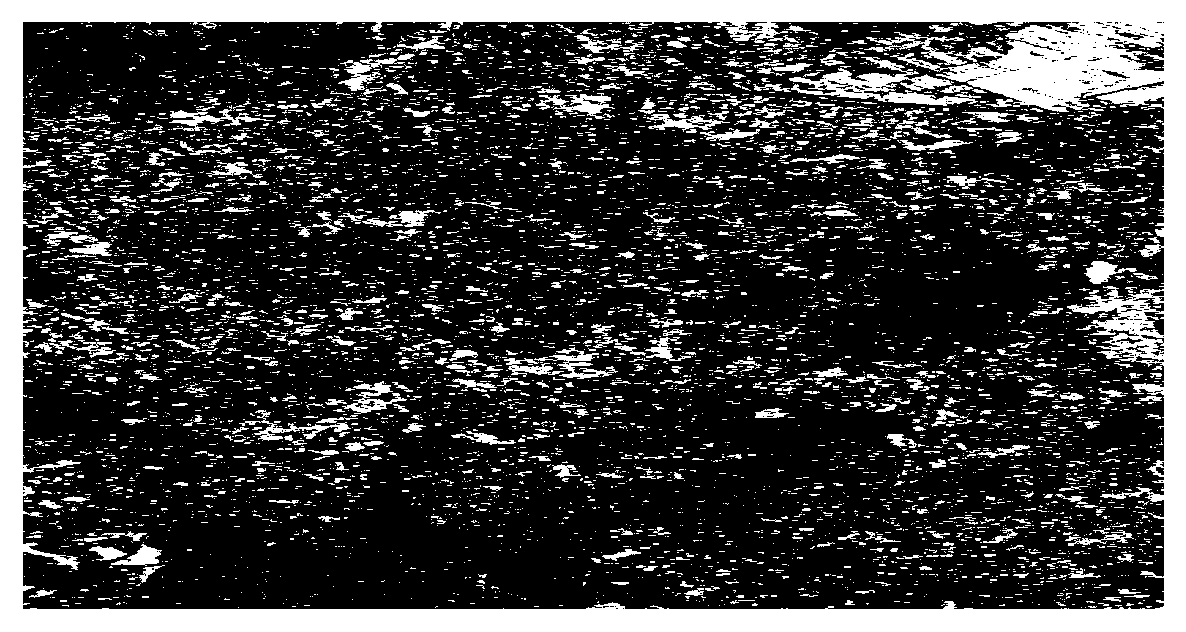}} \\
		\subfloat[Reference image.]{\includegraphics{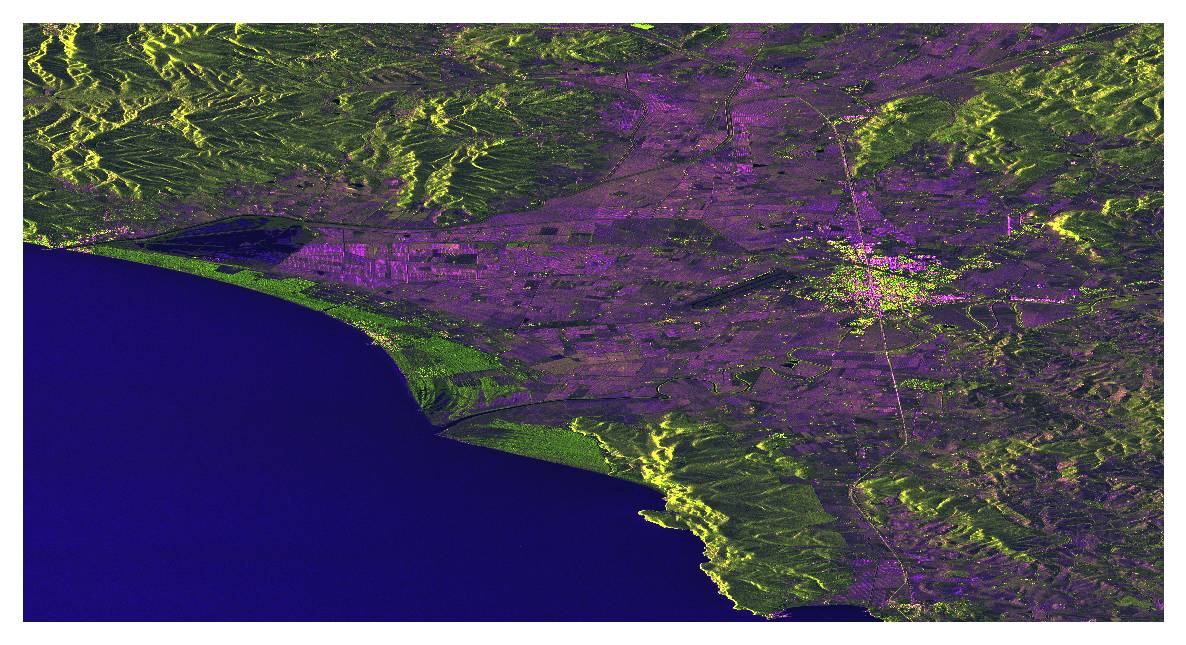}} 
		\subfloat[Change detection mask.]{\includegraphics{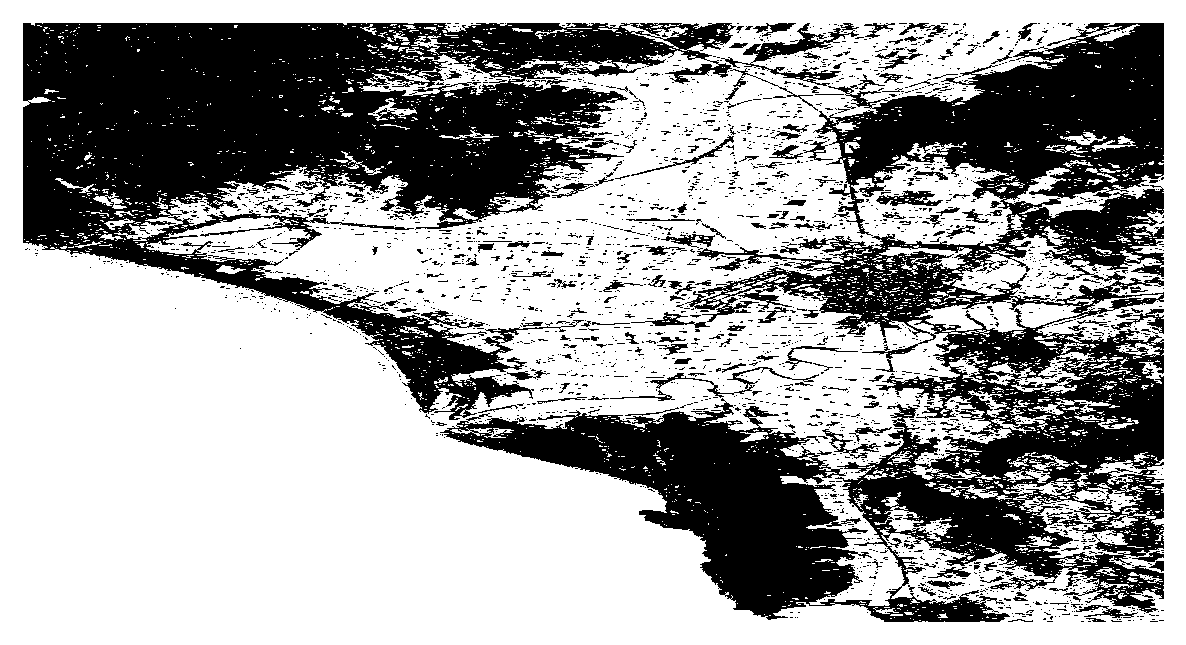}}

		\caption{Omnibus change detection test applied to an excerpt of the datasets over Murcia (top row) and Toscana (orbit 117)  (bottom row). }
		\label{fig:chg_masks}
	\end{figure*}

	For each time series, a change detection mask based on the omnibus statistical test explained in Subsection~\ref{ssec:cd_theory} is generated. A 4$\times$19 multilook filter, which yields a Equivalent Number of Looks (ENL) of around 40 has been used. Moreover, since the time series are long, and changes in the scene are more likely to happen, we set a low significance level of $10^{-10}$ to threshold the change detection probability shown in Eq.~\ref{eq:plnQ}. For instance, Figure~\ref{fig:chg_masks} shows the change detection masks of an excerpt of the scenes corresponding to the time series of Murcia and Toscana (orbit 117). As it can be observed by looking at Figs.~\ref{fig:chg_masks}(b) and (d), areas that remained unchanged during the time span covered by the data mainly correspond to urban and rural areas (black pixels in the masks), whereas agricultural and sea zones exhibit noticeable changes that are properly detected by the statistical test (white pixels).

	To generate the training dataset, a huge number of noisy--clean patches are randomly selected from each time series. The selection is random in terms of its location inside the image (spatially) and of its date in the whole time series (temporally). However, we only include patches that present almost no changes. More specifically, a patch is included in the training dataset only if less than 10\% of the pixels are detected as changed by the statistical test, otherwise it is discarded. 
 
 Unlike other works that use a larger patch size of 256$\times$256 pixels~\cite{Dalsasso2021, Tucker2022,Sebastianelli2022} to train the neural network, we prefer to employ a smaller block size of 64$\times$64 pixels. This allows us to ensure that, within an image patch, only a small amount of pixels, or even no pixels, present temporal changes. Therefore, intensity differences between the reference and noisy patches are solely due to the speckle component, rather than a combination of speckle and intensity variations caused by scene changes along time. This should prevent the convolutional neural network from learning erroneous features and predict biased image values when applied to new/unseen images. Finally, taking this into account, we can generate a vast number of noisy-clean patches from which the neural network can properly learn the underlying speckle component with a conventional supervised approach.
	
	\subsection{Filtering Performance Evaluation}
	\label{ssec:f_qual}
	
	The performance of the proposed method is evaluated in different ways. Firstly, with conventional indicators of the capability of the filter in terms of both speckle reduction and resolution preservation. Then, we also evaluate how the filter performs in rapidly-changing areas such as crops. This allows us to test whether the problem of temporal changes is correctly addressed by including the change detection strategy, so that the filter only removes speckle without altering the data or causing the appearance of artifacts. Finally, since the filter is applied on polarimetric SAR images, we include a typical step in polarimetric studies which is an example of a polarimetric decomposition.
	
	\subsubsection{Speckle Reduction and Resolution Preservation}
	
The Equivalent Number of Looks (ENL), computed over a manually-selected homogeneous region of interest $R$, provides an indicator of the smoothing capability of the method. Higher ENL values indicate a stronger speckle reduction capability. As described in~\cite{Anfinsen2009}, the ENL, adapted to the polarimetric case, can be calculated as:
	\begin{equation}
		\text{ENL} = \frac{[\tr(\mathbf{\bar{C}_f})]^2}{\tr(\mathbf{C_f}\mathbf{C_f})  - \tr(\mathbf{\bar{C}_f} \mathbf{\bar{C}_f} ) }
	\end{equation}
	where $\mathbf{C}_f$ is the filtered covariance matrix, and $\mathbf{\bar{C}}_f$ is the spatial average of the filtered covariance in an homogeneous area.
 
The resolution preservation capability of the filter can be assessed with the Edge Preservation Degree based on the Ratio of Average (EPD-ROA)~\cite{Feng2011_EPDROA}, which is similar to the conventional Edge Preservation Index (EPI)~\cite{DiMartino2014}, but better adapted to the multiplicative model of speckle. The value of the EPD-ROA can be obtained with dual-pol data by using the image span:
	\begin{equation}
		\text{EPD-ROA} = {\sum\limits_{i = 1}^{N}\left\vert \text{SPAN}_{f1}(i)/\text{SPAN}_{f2}(i)\right\vert\over \sum\limits_{i = 1}^{N}\left\vert \text{SPAN}_{o1}(i)/\text{SPAN}_{o2}(i)\right\vert}
	\end{equation}
	where $\text{SPAN}_{f1}$ and $\text{SPAN}_{f2}$ represent the span of adjacent pixel values of the filtered image along a given direction (horizontal or vertical), $\text{SPAN}_{o1}$ and $\text{SPAN}_{o2}$ correspond to the span of adjacent pixel values of the original image, and $N$ is the number of samples of the region where the index is calculated.
	
	Lastly, another indicator of how the speckle filter preserves the spatial resolution is the Structural Similarity Index (SSIM), calculated as:
	\begin{equation}
		\text{SSIM}(\mathbf{X}, \mathbf{Y}) = \frac{(2\mu_X\mu_Y + C_1)(2\sigma_{XY} + C_2)}{(\mu_X^2 + \mu_Y^2 + C_1)(\sigma_X^2 + \sigma_Y^2 + C_2)}
	\end{equation}
	where $\mathbf{X}$ and $\mathbf{Y}$ are two images, $\mu_X$, $\sigma_X$, $\mu_Y$, $\sigma_Y$ are the local means and variances computed for images $\mathbf{X}$ and $\mathbf{Y}$ respectively, and $C_1$ and $C_2$ are two constants included to avoid numerical instabilities if $(\mu_X + \mu_X)^2$ or $(\sigma_X + \sigma_X)^2$ are very small. Usually, $C_1 = 0.01L$ and $C_2 = 0.03L$, being $L$ the dynamic range of the data. In our case, we will calculate the SSIM of the span image (i.e., VV + VH intensity). Unlike the previous indicators, the SSIM requires a reference/ground truth to be calculated. In our case, we will use a temporal average of a stack of images as a reference, and manually select an urban area where more structures are present and changes are less likely to happen. 
	
	\subsubsection{Preservation of Temporal Patterns}
	
	The effective exploitation of multi-temporal stacks of images is a key aspect of the proposed method since a temporal average is used as reference to train the neural network model. However, as previously stated, the temporal average cannot be directly used if there are changes between acquisitions that might cause the neural network to learn erroneous features. To evaluate whether the inclusion of the change detection strategy benefits the learning process, we have carried out an experiment in which a whole time series of Sentinel-1 images corresponding to an agricultural area is processed with the neural network. Four crop fields are manually selected, and the temporal evolutions of both VV and VH intensities are analyzed. Since strong temporal changes are likely to happen in agricultural areas, we use them to test whether the filter introduces a bias in the resulting pixel values. Filtered intensities are compared with a spatial average of the original image data, since the spatial average of a completely homogeneous area provides the maximum likelihood estimator of the true intensity value for the whole area. Consequently, it is expected that the proposed speckle filter should provide an average intensity value close to that spatial average of the whole field (evidently, we assume that the crop is completely homogeneous and all the plants inside the crop have the same temporal evolution). It is worth mentioning that we do not have found such temporal analysis in other works that employed deep learning to despeckle SAR images, but this analysis constitutes a solid test to evaluate that the neural network was properly trained and does not produce biased filtered image values in rapidly-changing areas. 
	

	\subsubsection{Polarimetric Processing}
	
	A typical way to exploit polarimetric data is by means of target decompositions~\cite{cloudeBookPolarisation2009, lee2009polarimetric}. In this work, provided that we are using dual-pol images from Sentinel-1, we make use of a recently proposed model-based decomposition tailored to this type of data~\cite{Mascolo2022_decomp}. This decomposition provides two scattering components corresponding to a volume and a polarized scattering remainder. We will show an example of the decomposition output obtained after filtering the images with the proposed method based on a convolutional neural network, so as to show whether the filter provides the expected results.
	
	\subsubsection{Testing Images}
 \label{s:trainingimages}
	The evaluation of the speckle filtering proposed in this work has been carried out using Sentinel-1 images that were not used during the training phase of the neural network. Each quality metric is computed on a different dataset in order to highlight specific features (e.g., resolution preservation, despeckling strength, etc.). The details of the testing images are summarized in Table~\ref{t:dataset_test}.
	\begin{table}[ht!]
		
		\centering
		\caption{Characteristics of Sentinel-1 images used for testing.}
		\begin{tabular}{@{}lccccc@{}}
			\toprule
			Geographic area & Orbit (type)  & \thead{Acquisition dates} \\
			\toprule
			Munich (Germany)  & 168 (Descending) & 2021-01-19   \\
			San Francisco (USA) & 115 (Descending) & 2021-01-21 \\
			Barcelona (Spain) & 132 (Ascending)  &  2016-12-02 to 2020-12-29 \\
			Sevilla (Spain) & 74 (Ascending)  &  2022-03-26 to 2023-04-26  \\
			\bottomrule
		\end{tabular}
		\label{t:dataset_test}
	\end{table}

	\section{Results}
	\label{sec:results}
	
	The three image stacks of Murcia and Toscana (both orbits) are jointly processed. From them, a total of 140000 randomly-selected noisy--clean patches, with dimensions 64$\times$64 pixels and containing the four intensity bands defined in~(\ref{eq:C2_mod}), compose the training dataset. 
 The residual DnCNN neural network is trained for 140 epochs with a batch size of 32 samples and using the Adam optimizer with an initial learning rate of 0.001. The learning rate is reduced by a factor of 10 every 20 epochs. The neural network is trained on a NVIDIA DGX platform which has a NVIDIA A-100 GPU with 40 GB of VRAM and 6912 CUDA cores, and a Dual AMD Rome 7742 processor with 128 cores with a clock frequency of 2.25 GHz. The training of the neural network took 2 hours and 49 minutes.
	
	As it is stated in Section~\ref{s:trainingimages}, all the results shown in this section are obtained with Sentinel-1 images that were not used during the training phase. Despeckling performance of the proposed method is compared with two state-of-the-art polarimetric speckle filters which are not based on deep learning. Specifically, the NL-SAR~\cite{Deledalle2015a} and the MuLoG with BM3D~\cite{Deledalle_MuLoG} are used for comparison purposes, and also a conventional 4$\times$19 boxcar filter is employed as a baseline method. We have employed the official implementation of MuLoG available in: \url{https://pypi.org/project/mulog/}, and also the official distribution of the NL-SAR filter which can be found at: \url{https://www.charles-deledalle.fr/pages/nlsar.php}, with all parameters of each method set to default applying a single iteration.

		\subsection{Filtering Performance}
	\label{ss:indices}
 
\begin{figure*}
		\centering
		\setkeys{Gin}{width=0.5\linewidth}
		\subfloat[Original.]{\includegraphics{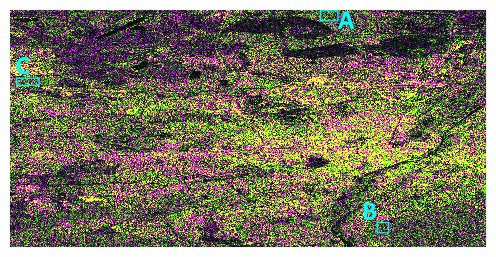}} 
		\subfloat[Boxcar.]{\includegraphics{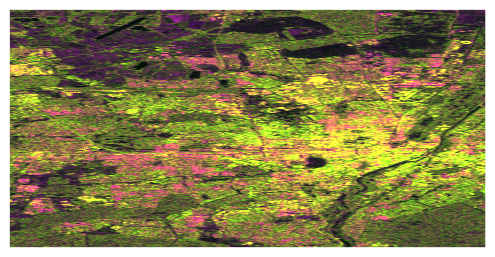}} \\ 
		\subfloat[NL-SAR.]{\includegraphics{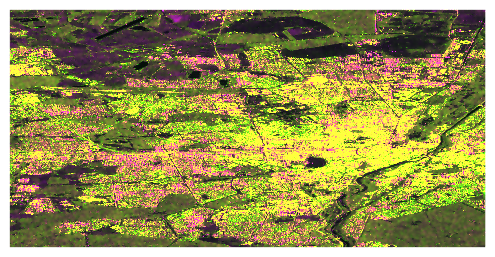}} 
		\subfloat[MuLoG.]{\includegraphics{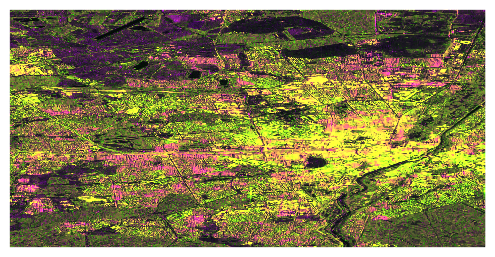}} \\
		\subfloat[Proposed.]{\includegraphics{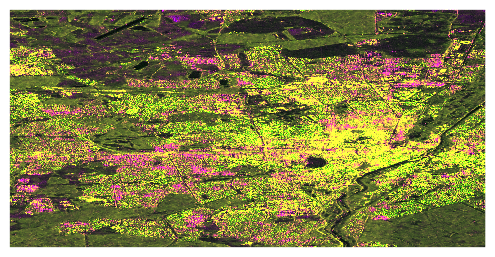}} 
		
		\caption{Filtering results of a Sentinel-1 image acquired over the city of Munich (Germany). Image dimensions are 1500$\times$3000 pixels in azimuth and range, respectively. Representation: RGB false colour composite formed with the backscattering coefficient at the linear channels with $R = VV$, $G = VH$, $B = VV/VH$.}
		\label{fig:munich}
	\end{figure*}

	Figure~\ref{fig:munich} shows the filtering results of a region of a Sentinel-1 dual-pol image acquired around the city of Munich (Germany) at date 2021-01-19. From a visual inspection of Figs.~\ref{fig:munich}(c),~(d) and~(e), we observe that the MuLoG, the NL-SAR and the proposed CNN-based filters present globally good results in terms of both speckle reduction and resolution preservation, whereas the boxcar filter (Fig.~\ref{fig:munich}(b)) blurs details of the image and clearly degrades its spatial resolution. Concerning the proposed method, it is worth mentioning that is does not seem to generate artifacts or \emph{invented} image values, so that the polarimetric radar response is preserved, independently of the scene type (i.e., in either high or low backscattering areas), as represented with the false-color RGB image of Fig.~\ref{fig:munich}(e).
 By comparing Figs.~\ref{fig:munich}(c) and (d), it can be seen that both the MuLoG and the NL-SAR adequately smooth homogeneous areas without degrading too much the original spatial resolution. Among these two filters, the NL-SAR seems to have a stronger smoothing capability, whereas the MuLoG preserves better fine structures and image textures (for instance, all the roads can be better visualized).

	Quantitative results of the performance of each method are shown in Table~\ref{t:ENL_munich}. The ENL is calculated in three different manually-selected homogeneous areas, named as A, B and C and shown in Fig.~\ref{fig:munich}(a). Even though it is difficult to find large and completely homogeneous areas, all methods (including also the boxcar filter), significantly increase the ENL in all selected areas. The NL-SAR exhibits the higher ENL in all areas, showing that it offers a stronger smoothing capability than the MuLoG and the proposed method, but oversmoothing effects are also visible for the NL-SAR. The proposed method effectively increases the ENL, showing a significant speckle reduction capability in homogeneous areas.

    \begin{table}[ht!]
    	\centering
   	    	\caption{Equivalent Number of Looks (ENL) calculated in three different Regions of Interest (RoI) of the Sentinel-1 image acquired over the city of Munich.}
    	\label{t:ENL_munich}
    	\begin{tabular}{@{}lcccccc@{}}
    		\toprule
    		{} & {} & \textbf{Equivalent Number of looks}  & {} \\
    		\toprule
    		{} & \textbf{RoI A} & \textbf{RoI B}  & \textbf{RoI C} \\
    		\toprule
    		Boxcar       & 49.4          &  58.3         & 77.8 \\
    		NL-SAR       & 89.7          & 125.6       & 145.8\\
		    MuLoG        & 66.9           & 90.2       & 95.5\\
    		\textbf{Proposed method} &  81.7  & 108.9 & 117.4 \\
    		\bottomrule
    	\end{tabular}
    \end{table}

\begin{figure*}
    	\centering
    	\setkeys{Gin}{width=0.5\linewidth}
    	\subfloat[Original.]{\includegraphics{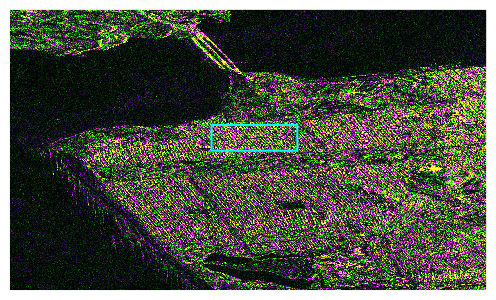}} 
    	\subfloat[Boxcar.]{\includegraphics{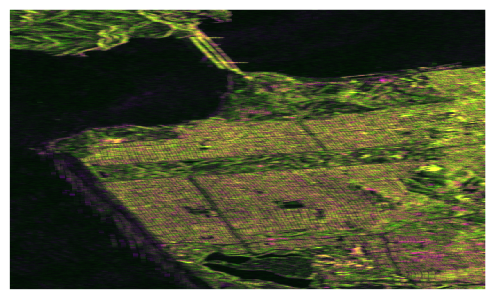}} \\ 
    	\subfloat[NL-SAR.]{\includegraphics{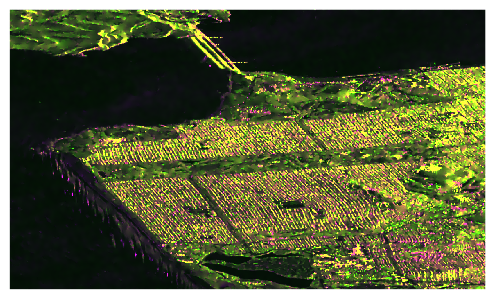}} 
    	\subfloat[MuLoG.]{\includegraphics{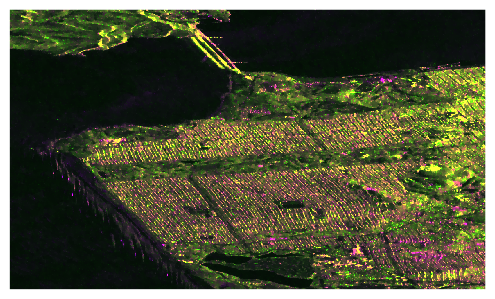}} \\
    	\subfloat[Proposed.]{\includegraphics{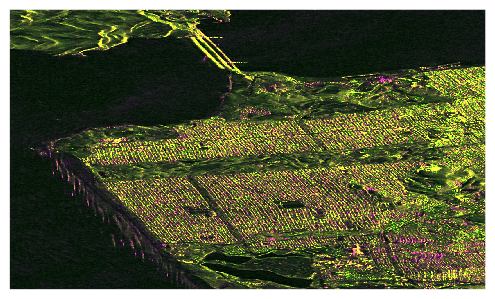}} 
    	
    	\caption{Filtering results of a Sentinel-1 image acquired over the city of Bay of San Francisco (USA).  Image dimensions are 1000$\times$2000 pixels in azimuth and range, respectively. Representation: RGB false colour composite formed with the backscattering coefficient  at the linear channels with $R = VV$, $G = VH$, $B = VV/VH$.}
    	\label{fig:sf}
    \end{figure*}

    The trained model is applied to another Sentinel-1 image acquired over the Bay of San Francisco (USA). The filtered polarimetric images are depicted in Fig.~\ref{fig:sf}. As in the previous example, the boxcar filter reduces speckle at the expense of a strong resolution loss, and the other three methods offer a better image restoration quality just by visual inspection. As shown in Fig.~\ref{fig:sf}(e), the proposed method retains many image details without blurring structures or other image features. Simultaneously, a significant smoothing is obtained over the sea area. It is worth mentioning that the neural network was not trained with image patches that covered sea areas, since they were detected as changing in the training scheme, but the fact that the filtered image is smoothed in such areas indicates again that the network is properly learning speckle characteristics.
    
    The quantitative assessment in terms of edge preservation is presented in Table~\ref{t:epd}, where the EPD-ROA is calculated in the area highlighted with the cyan rectangle in Fig.~\ref{fig:sf}(a), which corresponds to an urban area with presence of human-made structures. EPD values are calculated in the vertical and horizontal directions and are then averaged to provide a single value of the edge preservation capability.
    As expected, the boxcar filter, which averages heterogeneous zones, has the lowest edge preservation degree. The proposed method provides the best capability in preserving image details, with the MuLoG in second place. 

    \begin{table}[ht!]
    	\centering
    	\ra{1.2}
    	\caption{Edge Preservation Degree calculated in an urban area of the Sentinel-1 image acquired over the Bay of San Francisco.}
    	\label{t:epd}
    	\begin{tabular}{@{}lcccccc@{}}
    		\toprule
    		{} & \textbf{Edge Preservation Degree }  \\
    		\toprule
    		Boxcar       & 0.11             \\
    		NL-SAR       & 0.65             \\
 		    MuLoG        & 0.78             \\
    		\textbf{Proposed method} & 0.82 \\
    		\bottomrule
    	\end{tabular}
    \end{table}

\begin{figure*}
    	\centering
    	\setkeys{Gin}{width=0.48\linewidth}
    	\subfloat[Temporal average]{\includegraphics{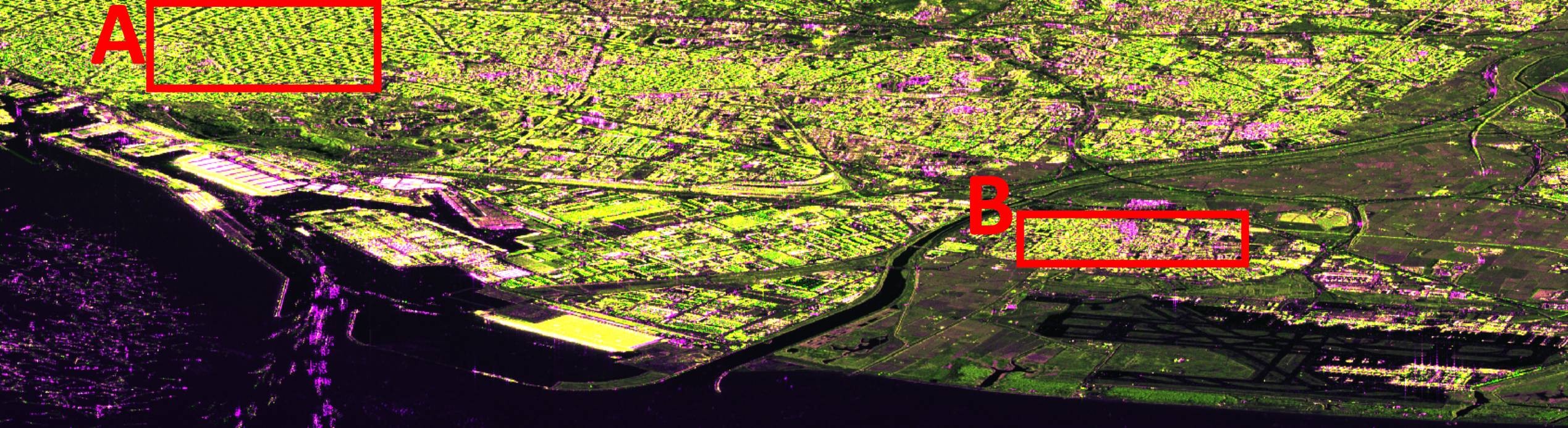}} \hspace{0.1em}
    	\subfloat[Original image acquired at 2017-08-17.]{\includegraphics{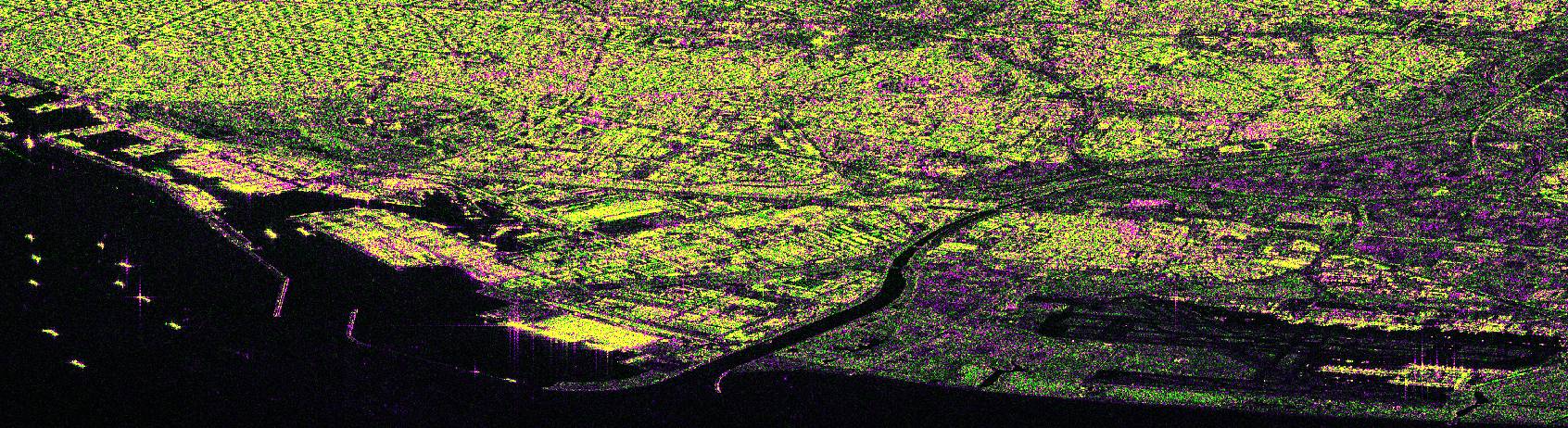}} \\
    	\subfloat[Boxcar.]{\includegraphics{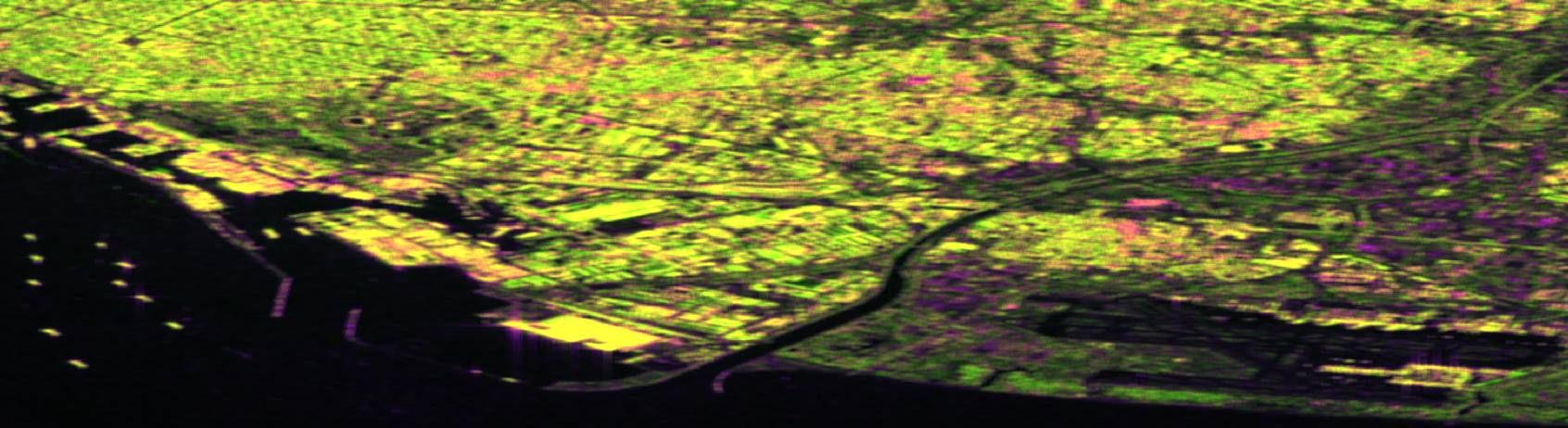}} \hspace{0.1em}
    	\subfloat[NL-SAR.]{\includegraphics{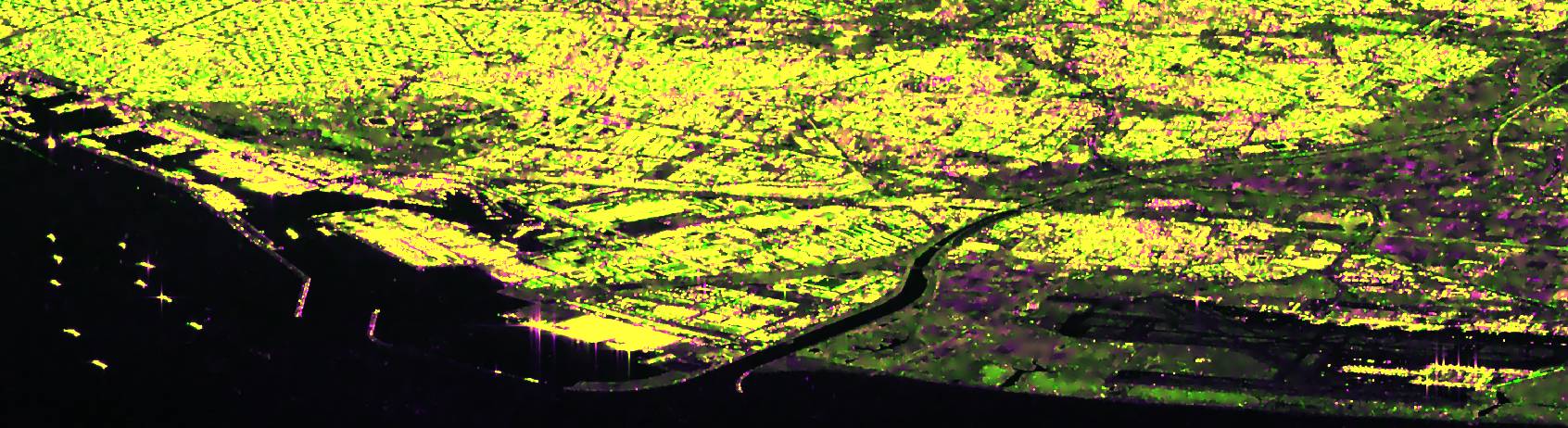}} \\
    	\subfloat[MuLoG.]{\includegraphics{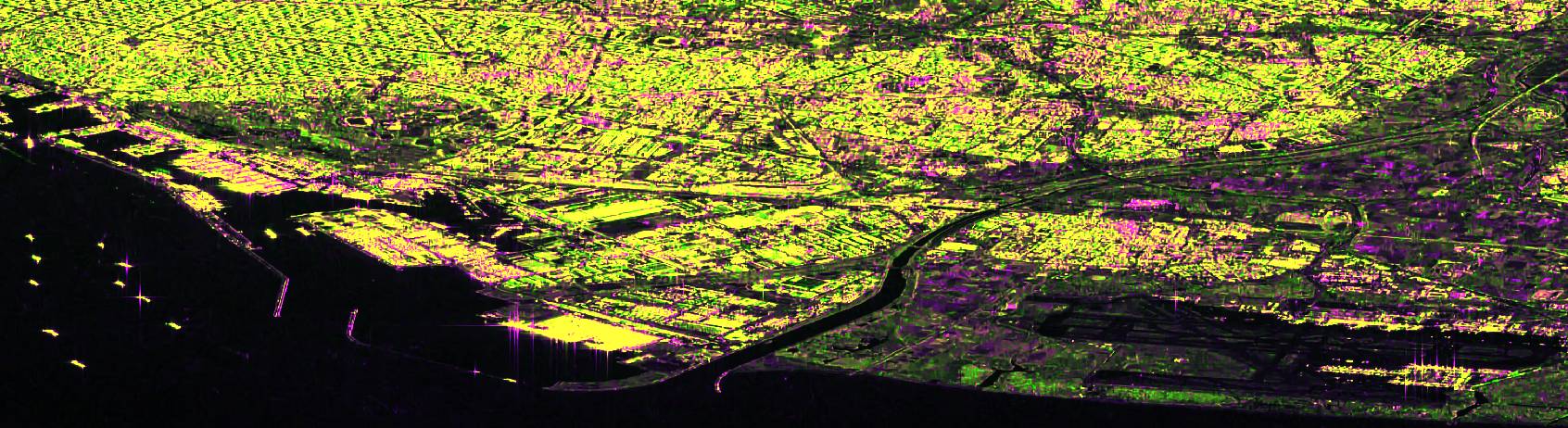}} \hspace{0.1em}
    	\subfloat[Proposed.]{\includegraphics{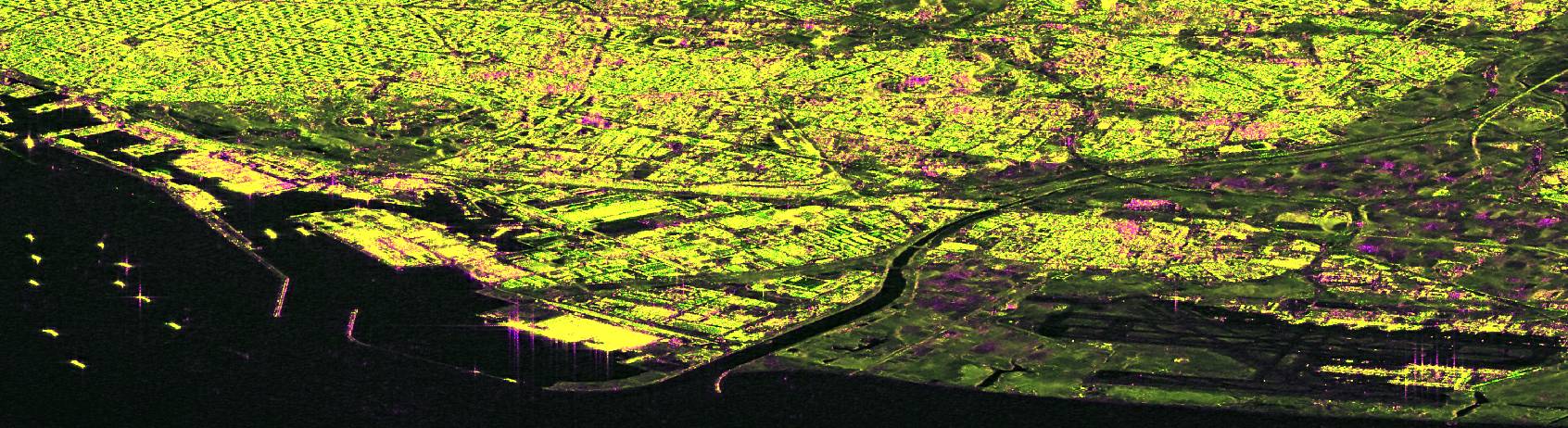}} 
    	
    	\caption{Filtering results of a Sentinel-1 image acquired over the city of Barcelona (Spain). Image dimensions are 1000$\times$4000 pixels in azimuth and range, respectively. Representation: RGB false colour composite formed with the backscattering coefficient at the linear channels with $R = VV$, $G = VH$, $B = VV/VH$.}
    	\label{f:barcelona}
    \end{figure*}

    The last quality estimator is the SSIM, widely used in image processing as an estimator of the similarity between noise-free and restored (i.e. filtered) images.
    Since SAR images without speckle cannot be obtained, we use as a reference image the temporally averaged stack of 120 coregistered images (acquired during years 2017 and 2018) and manually select different urban areas where human-made structures are present and which are less prone to changes. The area imaged is located in Barcelona (Spain). Figure~\ref{f:barcelona} shows the images we have employed to calculate the SSIM index. 
    
    Figure~\ref{f:barcelona}(a) corresponds to the temporal average. 
    Note the presence of artifacts at the bottom left part of Fig.~\ref{f:barcelona}(a). These artifacts are caused by bright points coming from vessels, which might be present or not in the images belonging to the time series, and lead to increased average values. This illustrates an extreme case of changes that we wanted to avoid when we generated the training dataset (as explained in Section~\ref{ssec:dataset}), since it would cause the neural network to learn erroneous features. Figure~\ref{f:barcelona}(b) shows an original SLC image, corresponding to a randomly-selected image from the stack and acquired at date 2017-08-17, and Figs.~\ref{f:barcelona}(c)-(f) display the filtering results obtained with each method. SSIM values are calculated on two different urban areas, labeled as A and B as shown in Fig.~\ref{f:barcelona}(a), and are summarized in Table~\ref{t:ssim_barcelona}. As expected, the Boxcar-filtered images have the lowest SSIM values as a consequence of the important degradation of the spatial resolution. The two state-of-the-art spatial-based methods, NL-SAR and MuLoG, exhibit higher SSIM indexes, showing again that are able to preserve fine image details and structures. Among them, the MuLoG filter offers higher SSIM values than the NL-SAR, probably because the latter presents a higher degree of overfiltering. Concerning the proposed method, it has a SSIM value very similar to that of MuLoG, which again shows a very remarkable resolution preservation capability.

    \begin{table}[ht!]
    	\centering
    	\ra{1.2}
    	\caption{Structural Similarity Index calculated in two urban areas of a Sentinel-1 image acquired over the city of Barcelona.}
    	\label{t:ssim_barcelona}
    	\begin{tabular}{@{}lcccccc@{}}
    		\toprule
    		{} & \textbf{Structural Similarity Index}  \\
    		\toprule
    		{} & \textbf{RoI A} & \textbf{RoI B}  \\
    		\toprule
    		Boxcar       & 0.16          &  0.09  \\
    		NL-SAR        & 0.35           & 0.24  \\
    		MuLoG       & 0.43          & 0.31  \\
    		\textbf{Proposed method} &  0.40 & 0.33      \\
    		\bottomrule
    	\end{tabular}
    \end{table}

	\subsection{Preservation of Temporal Patterns}
	\label{ss:temp_evo}

\begin{figure*}
		\centering
		\setkeys{Gin}{width=0.9\linewidth}
		\subfloat[Original.]{\includegraphics{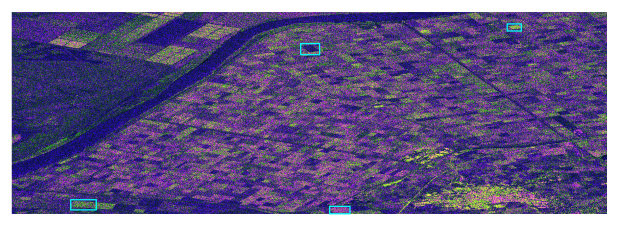}} \\ 
		\subfloat[Proposed method without change detection.]{\includegraphics{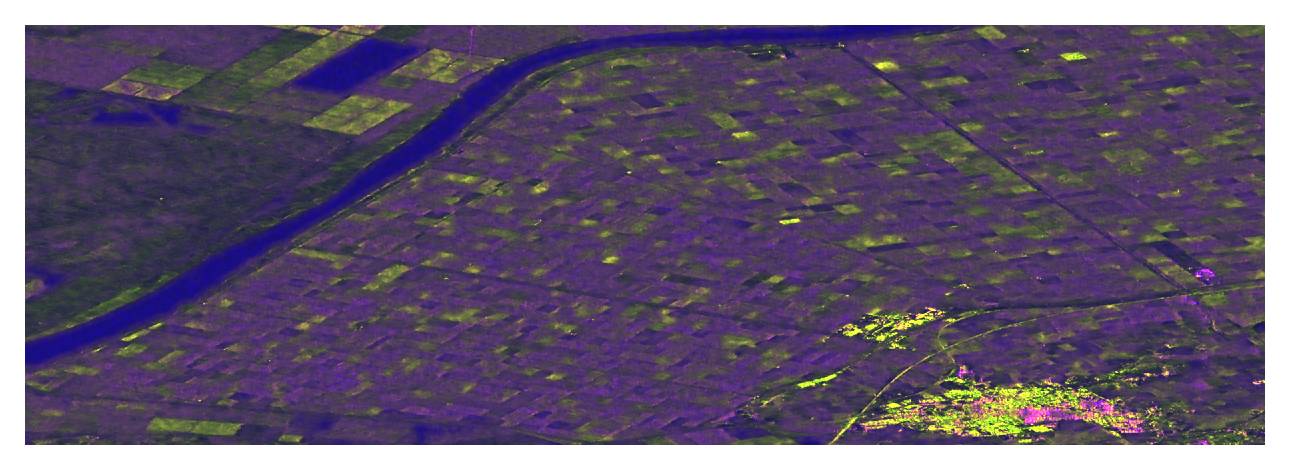}} \\
		\subfloat[Proposed method with change detection.]{\includegraphics{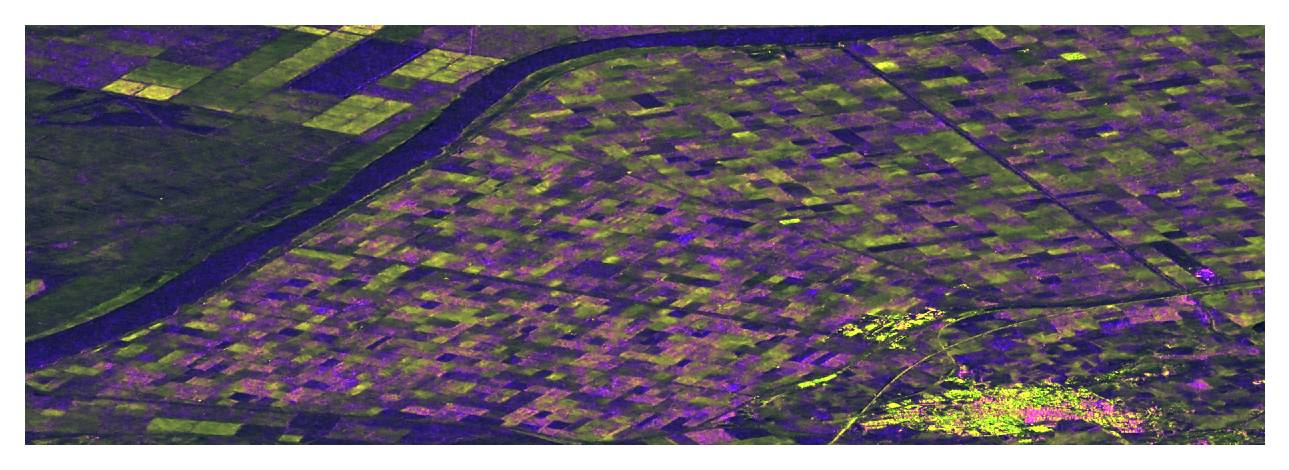}} \\

		\caption{Filtering results of a Sentinel-1 image acquired over an agricultural area in Sevilla province (Spain). Image acquired at date 2022-05-30.  Image dimensions are 1050$\times$3100 pixels in azimuth and range, respectively. Representation: RGB false colour composite formed with the backscattering coefficient  at the linear channels with $R = VV$, $G = VH$, $B = VV/VH$.}
		\label{f:sevilla}		
	\end{figure*}

	Figure~\ref{f:sevilla} shows an example of filtering results of one image of the time series acquired over Sevilla (Spain). As it can be seen, there are many crop fields with very different radar responses. By comparing Figs.~\ref{f:sevilla}(b) and (c), it can be deduced that both trained neural networks (with and without change detection) provide an important speckle reduction. However, when the neural network was trained without taking temporal changes into account (Fig.~\ref{f:sevilla}(b)), the resulting filtered image is clearly biased, and the polarimetric features (represented by colours) are significantly altered. In fact, filtered images seem to be biased towards a constant value, as shown by the very similar response that most crops have. Note that this effect is similar to the one shown in~\cite{Dalsasso2021}, when the SAR2SAR algorithm applied to single-pol data was trained without compensating temporal changes. Indeed, this bias in the filtered images is significantly reduced when temporal changes are considered for training the neural network, as shown in Fig.~\ref{f:sevilla}(c). 
	
\begin{figure*}[ht!]
        \captionsetup[subfigure]{labelformat=empty}
		\centering
		\setkeys{Gin}{width=0.45\linewidth}
        \subfloat[]{\includegraphics[width=5cm]{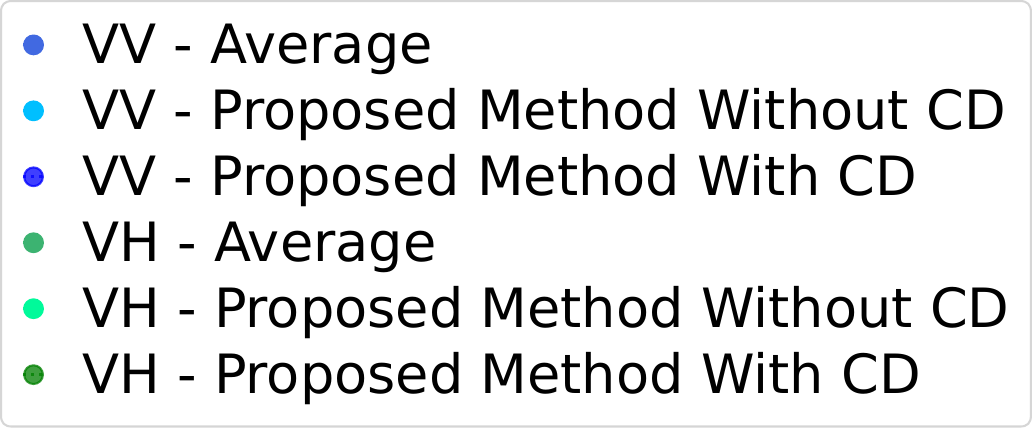}} \\
		\subfloat[(a)]{\includegraphics{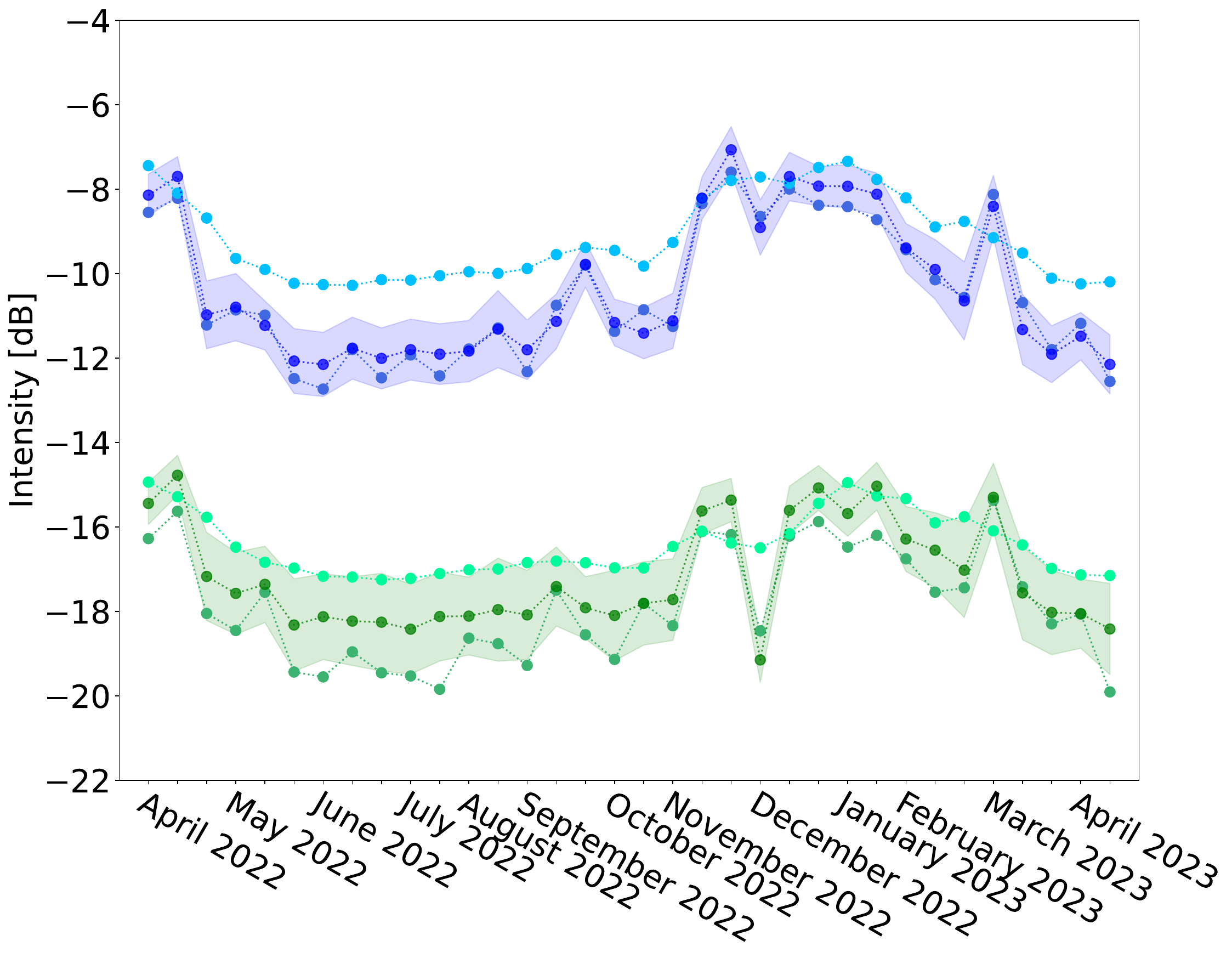}}
		\subfloat[(b)]{\includegraphics{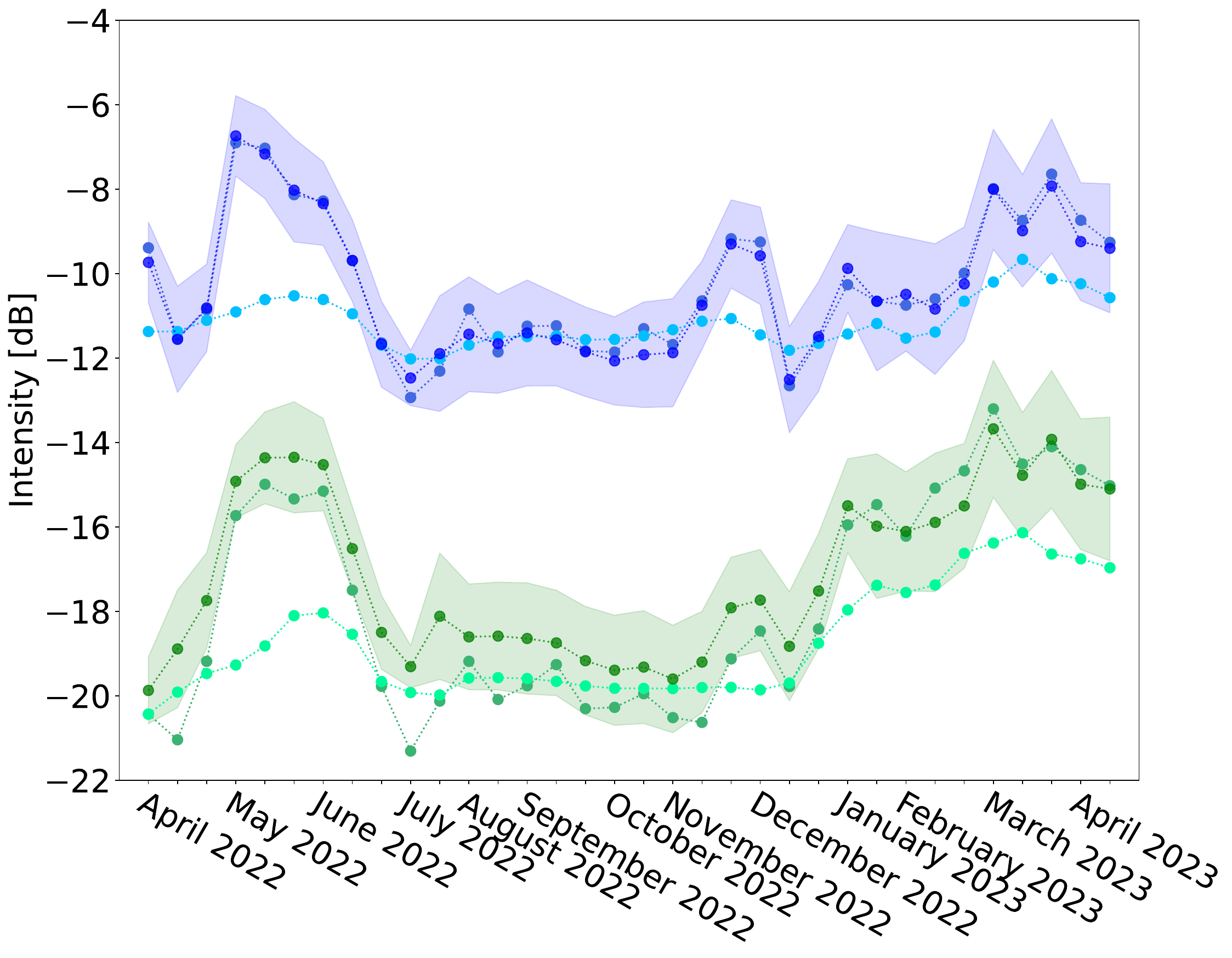}} \\
		\subfloat[(c)]{\includegraphics{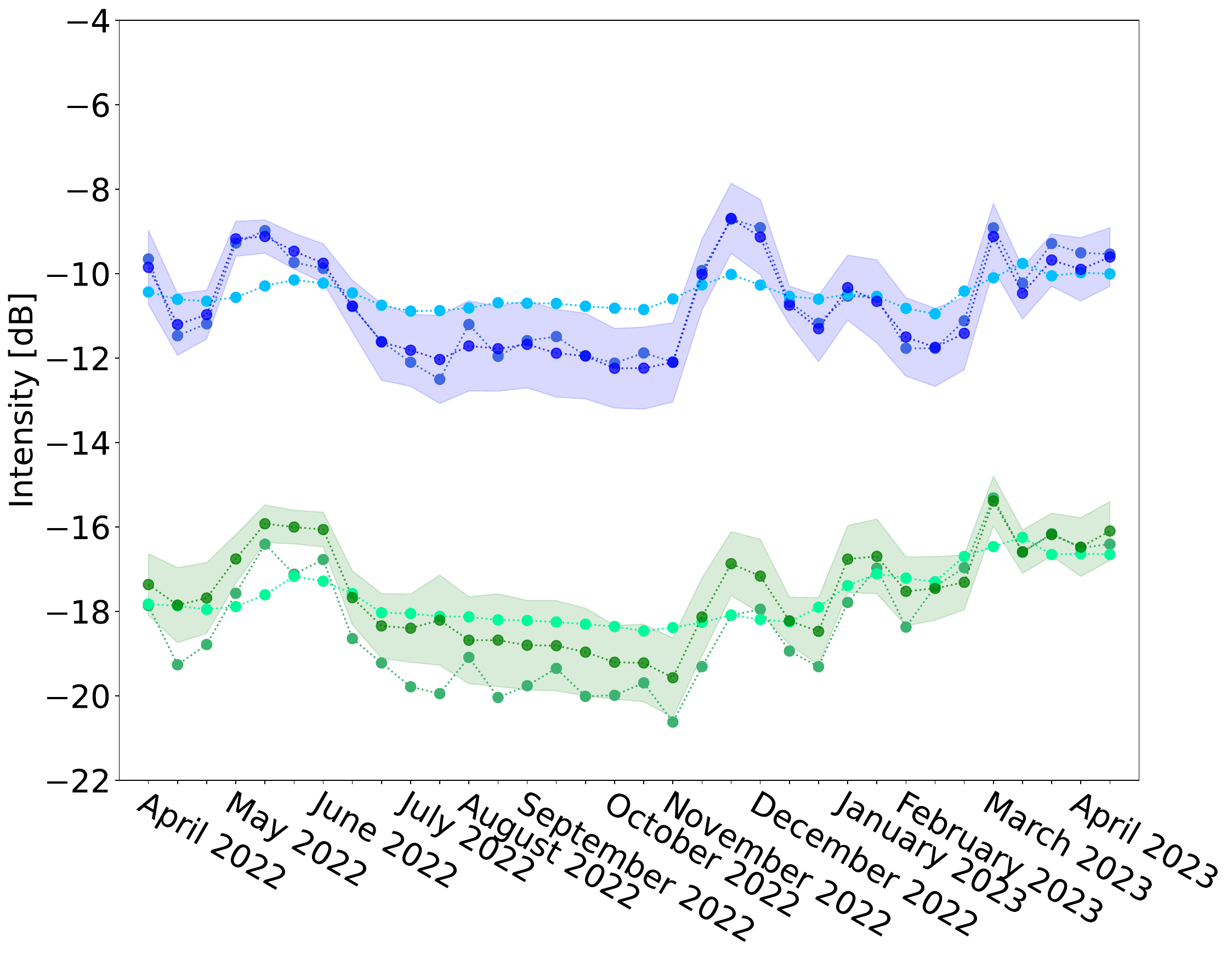}} 
		\subfloat[(d)]{\includegraphics{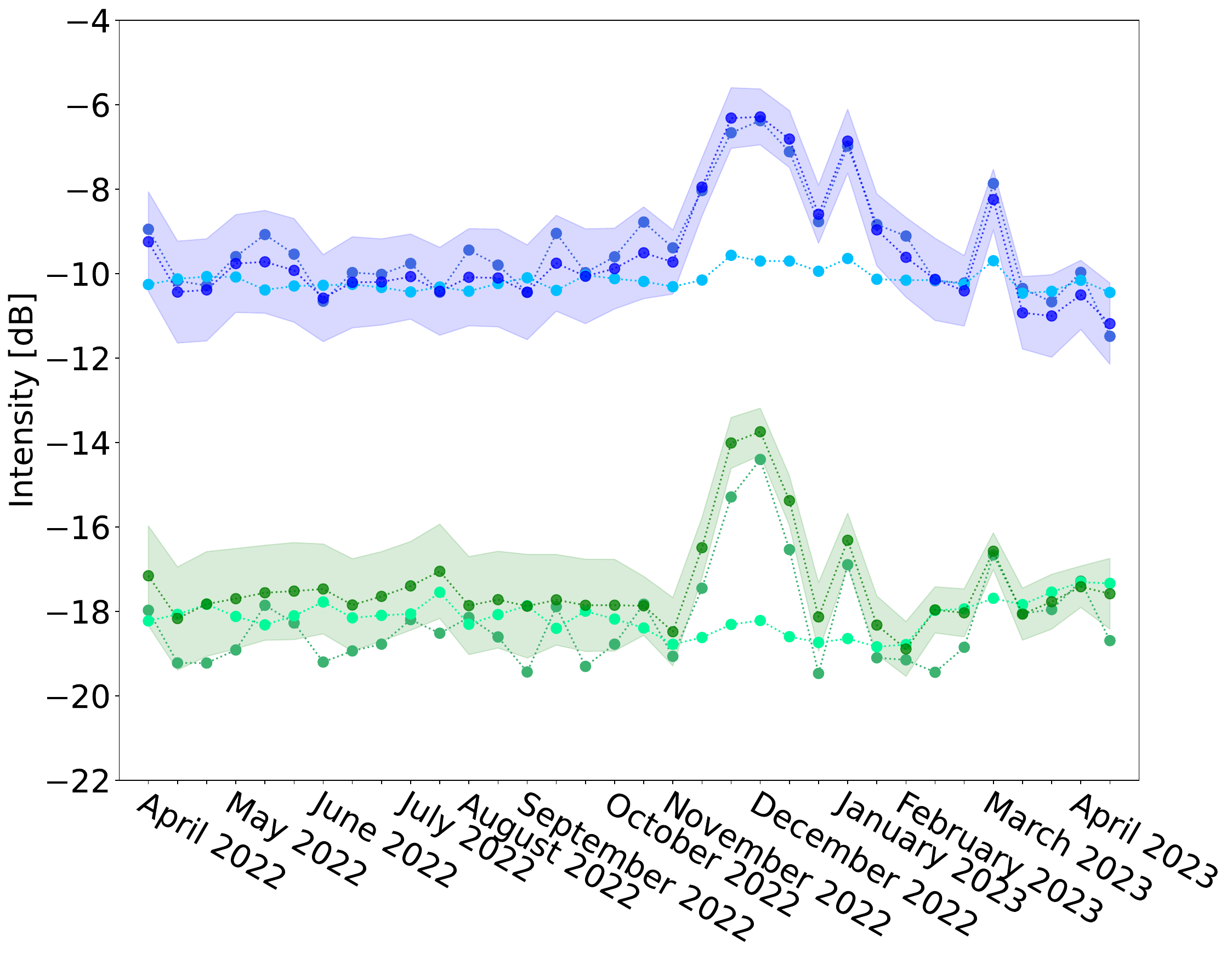}}
		
		\caption{Temporal evolution of VV and VH intensities associated with each selected field in the Sevilla test site. Solid lines denote the average value within the field, whereas the filled areas of the plots represent one standard deviation of the intensity values of the filtered data within the field.}
		\label{f:temp_evo}		
	\end{figure*}

	Besides the fact that, just by visual inspection, the filtered image looks unbiased and correctly restored with the proposed method, the following temporal analysis has been conducted to justify the importance of including the change detection strategy.
	We have manually selected four fields which are highlighted with the cyan rectangles shown in Fig.~\ref{f:sevilla}(a). For each date and for each field, the backscattering coefficient (or intensity) of both VV and VH channels are extracted, and the temporal evolution throughout the whole year of the unfiltered and filtered data with the proposed method are represented in Fig.~\ref{f:temp_evo}. 
	For comparison purposes, we also show the temporal evolution of the filtered images when the same neural network model was trained without taking temporal changes into account (i.e., when just a random selection of image patches is done, independently of the presence of changes). The number of pixels used to calculate the temporal statistics varies among fields, and lies between 625 and 1500. In any case, it is high enough to appropriately calculate the mean and standard deviation of the backscatter intensity within the field.
	 
	As illustrated in Fig.~\ref{f:temp_evo}, each field exhibits a unique temporal behavior due to the specific growth cycle of the crop type present in that field. Generally, the VV polarimetric channel shows a higher backscattering level compared to the VH channel. However, the key aspect we want to analyze here is the difference between the temporal evolution of the unfiltered and filtered data processed by the neural network. Firstly, it is clearly seen that there is a bias between the unfiltered and filtered data resulting from training the neural network without accounting for changes, as it could be predicted by looking at Figs.~\ref{f:sevilla}(a) and (b). In general, there is a significant difference, which goes up to more than 4 dB, between the filtered and the averaged original data, as shown for instance between May and July in Fig.~\ref{f:temp_evo}(a) and between November and January in Fig.~\ref{f:temp_evo}(d). More importantly, the main consequence of not accounting for changes is that the temporal evolution of crops is mostly "lost". That is, crop dynamical changes that cause temporal fluctuations in the intensities of VV and VH channels are not visible in the radar data. This can be clearly seen, for instance, between May and July in Fig.~\ref{f:temp_evo}(a) with an important increase of the backscatter intensity which is not visible in the filtered data, or with the drop of intensity that is shown in December in Fig.~\ref{f:temp_evo}(b). We deduce that when the neural network is trained without considering temporal changes, all potential changes along time are ignored, resulting in a biased learning towards the image mean (i.e., the constant value resulting from the temporal average). Since this average is not representative of a speckle-free image in areas with changes, the network is learning wrong features, and the speckle component cannot be removed properly. Consequently, temporal fluctuations of intensity are not tracked, and the neural network is always predicting a nearly constant value, as shown with all temporal evolutions in Fig.~\ref{f:temp_evo}, independently of the original intensity value.
	
	Conversely, when the change detection mask is included in the training of the neural network, the filtering results improve significantly. As shown in Figs.~\ref{f:temp_evo}(a)-(d), the difference, or bias, between the averaged unfiltered data and the filtered data is substantially reduced. In fact, the "true" intensity value, derived from averaging the original data, and the filtered data are consistently very close (less than 0.5 dB). In most cases, the difference falls in less than one standard deviation of the values of the filtered areas. This difference is smaller for the filtered VV polarization than for the VH channel. A greater difference is only observed for very low intensity values, close to the noise level (just below -20 dB) for the VH component, in which case the filtered data are slightly more biased towards higher backscattering values. 
	In addition, note that now the temporal evolutions of the filtered and the averaged original data are very similar, and all temporal fluctuations caused by the crop dynamics are now correctly obtained, proving the effectiveness of the change compensation strategy used to properly train the neural network.


	\subsection{Polarimetric Processing}
	
\begin{figure*}[ht!]
		\centering
		\setkeys{Gin}{width=0.9\linewidth} 
		\subfloat[Decomposition after filtering the data with the proposed method.] {\includegraphics{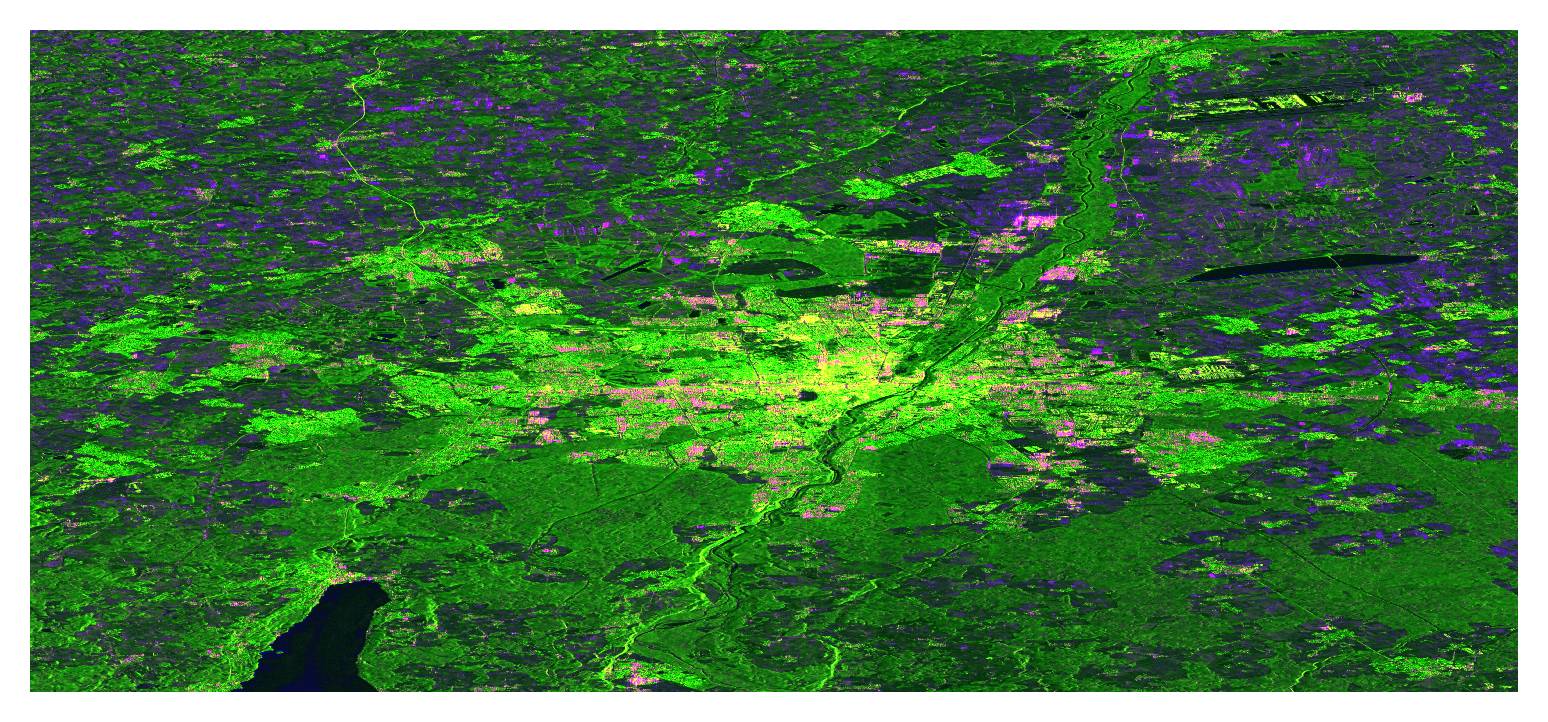}} \\
		\subfloat[Decomposition after filtering the data with a 4$\times$19 boxcar.]  {\includegraphics{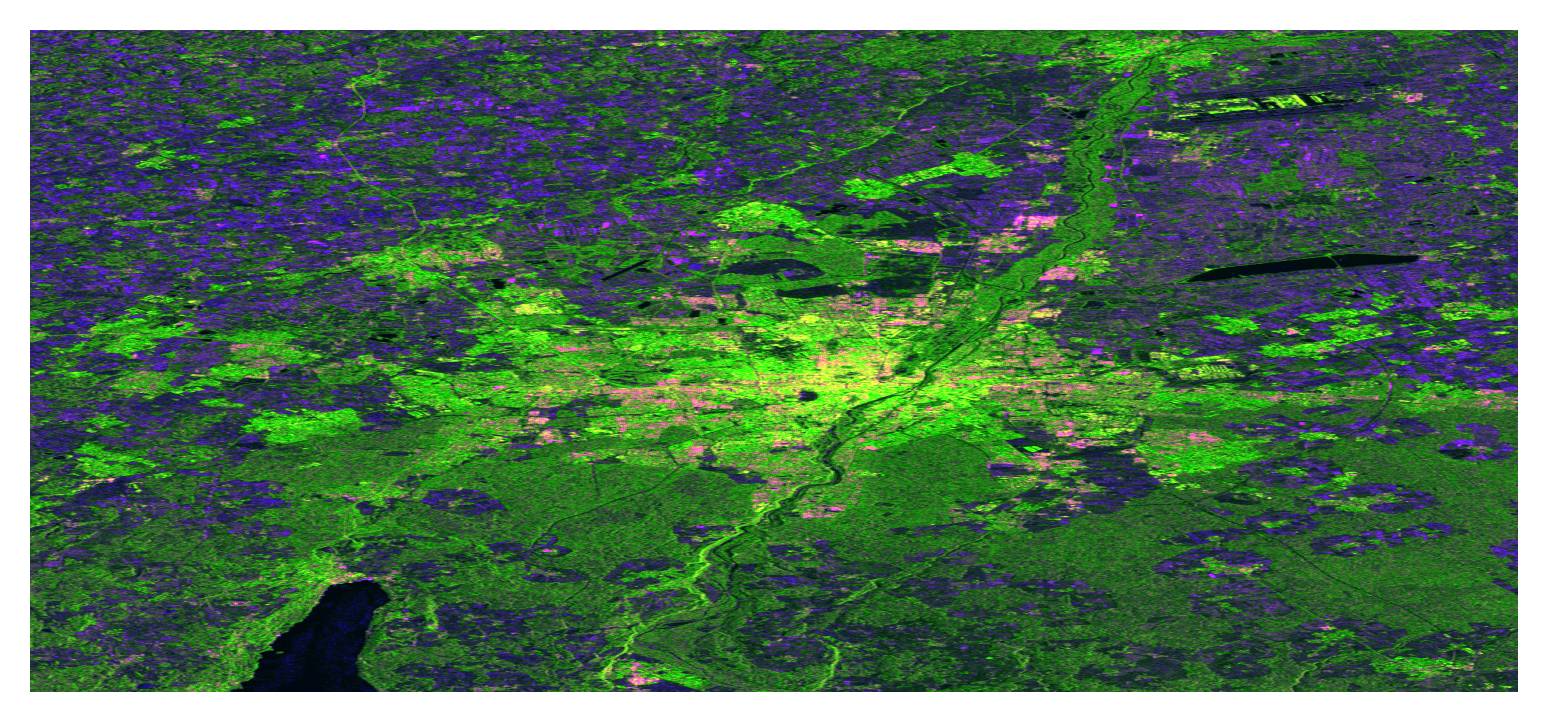}}

		\caption{Results of the polarimetric decomposition of the Sentinel-1 image acquired over the city of Munich: Representation: RGB false colour composite with R = $m_s$, G = $m_v$, and B = $m_s/m_v$.}
		\label{f:munich_decomp}

	\end{figure*}

	The model-based decomposition for dual-pol data proposed in~\cite{Mascolo2022_decomp} is applied to the filtered image of Munich, but a larger scene of 4500$\times$10000 pixels (azimuth$\times$range) that now includes the city of Munich and its surroundings is selected for a better visualization of the decomposition result in different land-cover types. The result of the decomposition can be visualized with the false-color RGB composite shown in Fig.~\ref{f:munich_decomp}. For comparison purposes, we also include the result obtained when the covariance matrix is filtered with a conventional 4$\times$19 boxcar.

By comparing Figs.~\ref{f:munich_decomp}(a) and (b), it can be seen that the same color pattern is obtained in both cases, which means that the proposed filter is not introducing artifacts or erroneous values. 
	Forest and vegetated areas are represented in green color, as a consequence of the dominant volume component $m_v$, whereas some agricultural areas are represented in a dark violet color (upper part of the images) which indicates that the polarized $m_s$ has a stronger response. Urban areas are represented in either pink or green colors, depending on their orientation with regard to the satellite. It is clearly appreciated that the result of the decomposition is much better when the covariance matrix was filtered with the proposed method. As shown in Fig.~\ref{f:munich_decomp}(a), details are much better preserved so that the decomposition yields more accurate results in all parts of the image. Note that, evidently, the analysis of the decomposition is out of the scope of this work, so that we only wanted to show an example of the advantage of filtering a dual-pol image with the proposed speckle filter.

	\subsection{Computational Cost}
	
	A complete and fair analysis of the computational cost cannot be done since the software implementation of each method is completely different. The NL-SAR filter is implemented in C language and includes multi-core and parallel processing, whereas the MuLoG method is implemented in Python language (which is a higher level programming language and therefore slower). The proposed method is implemented using \textit{TensorFlow-Keras} and takes advantage of GPU technology (like the vast majority of deep learning-based applications). 
	
	The main computational burden of both NL-SAR and MuLoG-BM3D corresponds to the cost of calculating similar patches or pixels to average with the pixel under test. Concerning the proposed method based on deep learning, the training part has the highest computational burden and requires GPU processing. As previously stated, the neural network was trained on high-performance computing (HPC) system to take advantage of GPU and CUDA, and took 2 hours and 49 mins. However, once the model is trained, a SAR image can be filtered using either GPU or CPU.
	
	Table~\ref{t:comp_times} provides the computation times required to filter the Sentinel-1 image of Munich shown in Fig.~\ref{fig:munich}, the dimensions of which are 1500~$\times$~3000 pixels. Computation times are obtained using a personal computer having an Intel Core i7-8700k CPU with a clock rate of 3.70 GHz, 64 GB of RAM, and a NVIDIA GTX 1080 (8 GB of VRAM). As detailed in Table~\ref{t:comp_times}, the NL-SAR has the fastest processing time and the Python version of MuLoG is the slowest. Concerning the proposed method, computation times obtained using CPU and GPU are very similar, which is expected since the prediction, that is, the application of the trained weights of the network on an image block of 64~$\times$~64 pixels is not computationally intensive. In this regard, note that when the same image is filtered on a HPC NVIDIA DGX platform (with a NVIDIA A-100 graphic card), the computation time is still very similar to the ones obtained using only the CPU or a low-performance GPU of a personal computer.

	\begin{table}[ht!]
		\centering
		\ra{1.2}
		\caption{Computation times for filtering the Sentinel-1 Image of Munich shown in Fig.~\ref{fig:munich} with different methods.}
		\label{t:comp_times}
		\begin{tabular}{@{}lcccccc@{}}
         \toprule
    		{} & \textbf{Computation Time (s)}  \\
            \toprule
			NL-SAR                       & 203       \\
			MuLoG                        & 1567      \\
			Proposed Method - CPU        &  553      \\
			Proposed Method - NVIDIA GTX 1080        &  517      \\
			Proposed Method - NVIDIA A-100    &  501      \\
			\bottomrule
		\end{tabular}
	\end{table}

	\section{Conclusions}
	\label{sec:concl}
	In this work, a complete framework to train a convolutional neural network with the goal of reducing speckle in polarimetric SAR images is proposed. The method exploits various multi-temporal stacks of co-registered SAR images, and a large amount of randomly-selected noisy--reference patches are used as input to the network, which is trained by following a supervised approach. The methodology includes a change detection strategy to avoid the neural network to learn erroneous image features. Specifically, the omnibus statistical test is used to detect temporal changes that happened in the time-span covered by the data, so that image patches of changed areas are not used as input to the neural network. We have shown that the inclusion of the change detection strategy greatly improves the filtering performance by reducing the bias, and, hence, the appearance of wrong pixel values or artifacts in the filtered images. This is because the inclusion of change detection allows to train the neural network with a vast amount of noisy--reference pairs of images, where the only difference is the underlying speckle component. Consequently, the residual model allows to properly estimate the speckle and remove it from the data. We have shown that the proposed method offers very remarkable results in terms of both speckle reduction and resolution preservation. In addition, the trained model can be properly used for further polarimetric processing with either one image or a multi-temporal stack of images. 
	
	As future work, we plan to improve the results by evaluating other neural network architectures or by changing the learning approach we have followed (i.e., from a supervised to semi-supervised or even an unsupervised strategy with polarimetric images). The application of the same methodology to quad-polarimetric data will be analysed by exploiting stacks of Radarsat-2 and ALOS-2 images. Finally, for further investigation and evaluation of the proposed methodology, the weights of the trained model, testing scripts and additional filtering results are released along with this paper and are made public in a Git repository\footnote{\url{https://github.com/alejandromestrequereda/CNN_PolSAR_Despeckling/}}.



	\bibliographystyle{IEEEtran}
	\bibliography{IEEEabrv,references}

	
	
		
	
	

\end{document}